\documentclass[final]{cvpr}

\usepackage{times}
\usepackage{epsfig}
\usepackage{graphicx}
\usepackage{amsmath}
\usepackage{amssymb}

\usepackage[utf8]{inputenc} %
\usepackage[T1]{fontenc}    %
\usepackage[backref=page]{hyperref}       %
\usepackage{url}            %
\usepackage{booktabs}       %
\usepackage{amsfonts}       %
\usepackage{nicefrac}       %
\usepackage{microtype}      %

\usepackage{color}
\usepackage{mathtools}
\usepackage{siunitx}
\usepackage{graphicx}
\usepackage{grffile}
\usepackage{comment}
\usepackage{multirow}
\usepackage{adjustbox}
\usepackage{caption}

\newcommand{\R}{\mathbb{R}}
\DeclarePairedDelimiter\abs{\lvert}{\rvert}%
\newcommand{\nm}[1]{
  \ifthenelse{\equal{#1}{}}{\norm{\cdot}}{\norm{#1}}
}
\definecolor{armygreen}{rgb}{0.0, 0.5, 0.0}
\definecolor{amber(sae/ece)}{rgb}{1.0, 0.49, 0.0}
\DeclarePairedDelimiterX{\norm}[1]{\lVert}{\rVert}{#1}

\newcommand{\sm}[1]{\textcolor{black}{#1}}
\newcommand{\shivam}[1]{\textcolor{black}{#1}}
\newcommand{\yun}[1]{\textcolor{black}{#1}}

\usepackage{bm}

\iftrue %
\newcommand{\cutsectionup}{\vspace*{-0pt}} 
\newcommand{\cutsectiondown}{\vspace*{-0pt}}

\newcommand{\cutsubsectionup}{\vspace*{-0pt}}
\newcommand{\cutsubsectiondown}{\vspace*{-0pt}}

\newcommand{\cutparagraphup}{\vspace*{-8.5pt}}

\newcommand{\cutcaptionup}{\vspace*{-7pt}}
\newcommand{\cutcaptiondown}{\vspace*{-4pt}}

\newcommand{\cuthalftablecaptiondown}{\vspace*{-2pt}}

\newcommand{\cuttablecaptionup}{\vspace*{-0pt}}
\newcommand{\cuttablecaptiondown}{\vspace*{-0pt}}

\newcommand{\cutabstractdown}{\vspace*{-0pt}}

\else %
\newcommand{\cutsectionup}{}
\newcommand{\cutsectiondown}{}

\newcommand{\cutsubsectionup}{}
\newcommand{\cutsubsectiondown}{}

\newcommand{\cutparagraphup}{}

\newcommand{\cutcaptionup}{}
\newcommand{\cutcaptiondown}{}

\newcommand{\cuthalftablecaptiondown}{}

\newcommand{\cuttablecaptionup}{}
\newcommand{\cuttablecaptiondown}{}

\newcommand{\cutabstractdown}{}

\fi

\newcommand{\shorttitle}{\textsc{GeoSim}}

\def\cvprPaperID{4877} %
\def\confYear{CVPR 2021}

\begin{document}
\thispagestyle{empty}

\title{GeoSim: Realistic Video Simulation via Geometry-Aware Composition for Self-Driving}

\author{
  Yun Chen$^{1}\thanks{Equal Contribution} $\quad Frieda Rong$^{1,3}\footnotemark[1]$ \quad Shivam Duggal$^{1}\footnotemark[1]$ \quad Shenlong Wang$^{1,2}$  \quad Xinchen Yan$^{1}$ \\ \quad Sivabalan Manivasagam$^{1,2}$  \quad Shangjie Xue$^{1,4}$  \quad Ersin Yumer$^{1}$ \quad Raquel Urtasun$^{1,2}$\\
 $^{1}$Uber Advanced Technologies Group \quad $^{2}$University of Toronto \\
 $^{3}$ Stanford University \quad $^{4}$Massachusetts Institute of Techonology  \\
 \small\texttt{\{chenyuntc, shivamduggal.9507, skywalkeryxc, meyumer\}@gmail.com,
  }\\
  \small\texttt{rongf@cs.stanford.edu},
  \small\texttt{\{slwang, manivasagam, urtasun\}@cs.toronto.edu, sjxue@mit.edu
  }
}

\maketitle

\begin{abstract}

   Scalable sensor simulation is an important yet challenging open problem for safety-critical domains such as self-driving. Current \yun{works}
   in image simulation either fail to be photorealistic or do not
   model the 3D environment and the dynamic objects within, losing high-level control and physical realism. In this paper, we present GeoSim, a geometry-aware image composition process which synthesizes novel urban driving 
   \sm{scenarios}
   by augmenting
   existing
   images with dynamic objects extracted from other scenes and rendered at
   novel
   poses.
   Towards this goal,
   we first build a diverse bank of 3D  objects with both realistic geometry and appearance from sensor data.
   During simulation,
   we perform a novel
   geometry-aware simulation-by-composition procedure
   which
   1) proposes plausible and realistic object placements into a given scene, 2) renders novel views of dynamic objects from the asset bank, and 3) composes and blends the rendered image segments. The resulting synthetic images are 
   realistic, \yun{traffic-aware}, and geometrically consistent, allowing 
   \sm{our approach}
   to scale to complex use cases. We demonstrate two such important applications:
   long-range realistic video simulation across multiple camera sensors, and synthetic data generation for data augmentation on downstream segmentation tasks.
   Please check
   \href{https://tmux.top/publication/geosim/}{\textcolor{blue}{https://tmux.top/publication/geosim/}} for high-resolution video results.
   
\end{abstract}
\cutabstractdown

\cutsectionup
\section{Introduction}
\cutsectiondown

\begin{figure*}[t]
    \includegraphics[width=\linewidth]{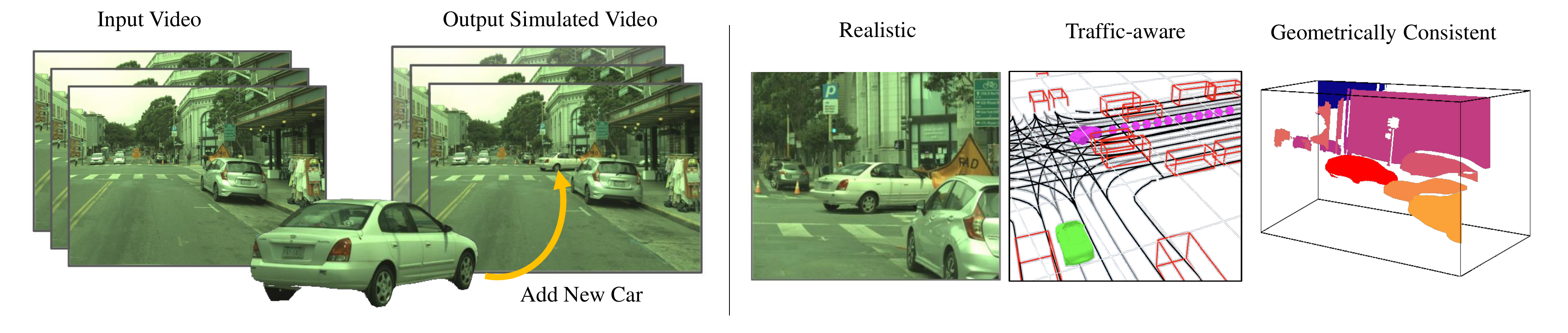}
    \cutcaptionup
    \vspace{-.5em}
    \caption{\textbf{Realistic video simulation via geometry-aware composition for self-driving.} We proposed a novel data-driven image manipulation approach that inserts dynamic objects into existing videos. Our resulting synthetic video footages are 
    \sm{highly realistic,}
    layout-aware, and geometrically consistent, allowing image simulation to scale to complex use cases.}
    \label{figure:pipeline}
    \cutcaptiondown
    \vspace{-.5em}
    \end{figure*}

Walking along an empty pavement on a silent Sunday morning, one can easily fantasize
how busy it could look during rush hour on a weekday, or how a parked car might look when driving \sm{on}
a different street. %
Humans are capable of recreating the experience of visually perceiving objects and scenes to generate new visual data in their minds.
Such an ability allows us to formulate novel scenarios and synthesize events in our heads without \sm{experiencing it directly.}

Researchers have devoted significant effort towards enhancing computers with the capability of creating
pictures by replicating visual content~\cite{tewari_state_2020}.
This brings immense value to many industries, such as film making, robot simulation, augmented reality, and teleconferencing.
In the literature, two main paradigms exist: \textit{computer graphics} approaches and \textit{image editing} methods.
Computer graphics models the image generation %
process \sm{with physics, by first }
creating a virtual 3D world and then \sm{mimicking}
how light is transmitted within the world to produce a realistic scene rendering.

To produce visually appealing 
results, physics-based rendering requires a significant amount of computing resources, costly manual asset creation, and physical modeling \cite{corona}. 
Images produced by existing real-time rendering engines~\cite{unreal, martinez2017beyond, dosovitskiy_carla_2017}, 
still have a significant realism gap, reducing their impact in robot simulation and data augmentation for training.
{Data-driven image editing} methods such as {image composition}~\cite{karsch2011rendering, lalonde2007photo, cong2020dovenet, dwibedi2017cut, bhattad2020cut}
and generative image synthesis \cite{wang2016generative,yang_high-resolution_2016,wang_high-resolution_2017,chen_photographic_2017,johnson2018image,park2019semantic} have received significant attention over the past few years. %
They focus on pushing realism through generative models trained from large-scale visual data.
However, most of the efforts do not correspond to an underlying realistic 3D world, and as a consequence, the generated 2D contents are not directly useful for applications such as  3D gaming and robot simulation.

In this paper, we propose  {\it GeoSim},  a realistic   image manipulation framework that inserts dynamic objects into existing videos.
GeoSim exploits a combination of both
data-driven approaches
and
computer graphics to \sm{generate assets inexpensively while maintaining high visual quality through}
physically grounded simulation.
In particular, by leveraging
low-cost bounding box annotations and sensor data
captured by a self-driving fleet driving
\sm{around} multiple U.S. cities,
GeoSim builds a fully-textured large-scale 3D assets bank. 
While self-driving data %
 is widely available ~\cite{kitti,argoverse, nutonomy, waymo}, it is non-trivial to automatically build these assets
due to the sparsity of the 3D observations, 
occlusions, shadows,  limited viewpoints,  and lighting changes. 
Our asset reconstruction is robust to these challenges, as we ensure consistency 
\sm{across} %
multiple observations in time and learn a strong shape prior to regularize our assets.
GeoSim then exploits the 3D scene layout (from high-definition (HD) maps and  LiDAR data)
to add vehicles in plausible locations
and make them behave realistically by considering the full scene.
Finally, using this new 3D scene, GeoSim performs 
image-based rendering to properly handle occlusions, and neural network-based image in-painting to ensure the inserted object seamlessly blends in by filling holes, adjusting color inconsistencies due to lighting changes,  and removing sharp boundaries.

Using GeoSim, our resulting synthetic images and video footages are 
\sm{realistic,}
dynamically plausible, and geometrically consistent.
We
showcase two important applications:
long-range realistic video simulation across multiple camera sensors
and synthetic labeled data generation for {training} 
self-driving perception algorithms.
Our approach outperforms prior work \sm{on}
both qualitative and quantitative realism metrics. 
We also see significant gains on perception performance when leveraging GeoSim images.
These experiments suggest the potential of GeoSim for a plethora of  applications,
such as realistic safety verification, data augmentation, Sim2Real, augmented reality, and automatic video editing.

\cutsectionup
\section{Related Work}
\cutsectiondown
\begin{figure*}[t]
\begin{center}
\begin{minipage}[t]{.61\linewidth}
\vspace{0pt}
\centering
\adjustimage{width=\linewidth,trim={0 },clip,valign=m}{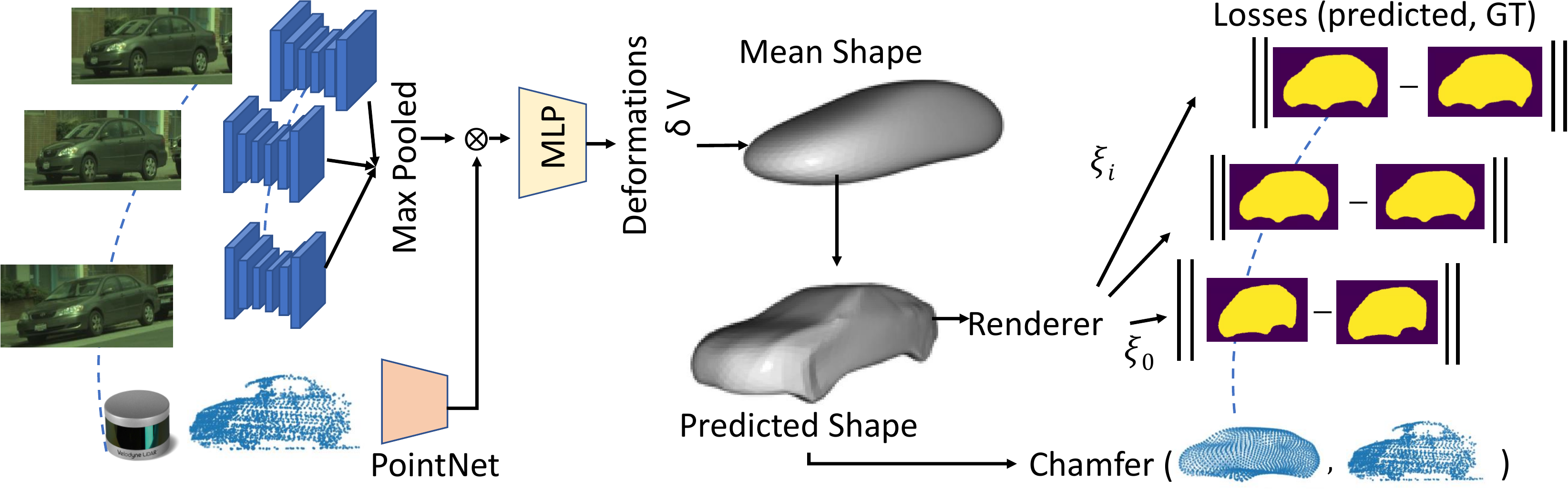}
\label{fig:mesh-reconstruction}
\end{minipage}%
\begin{minipage}[t]{.39\linewidth}
\vspace{0pt}
\centering
\adjustimage{width=\linewidth,trim={0 },clip,valign=m}{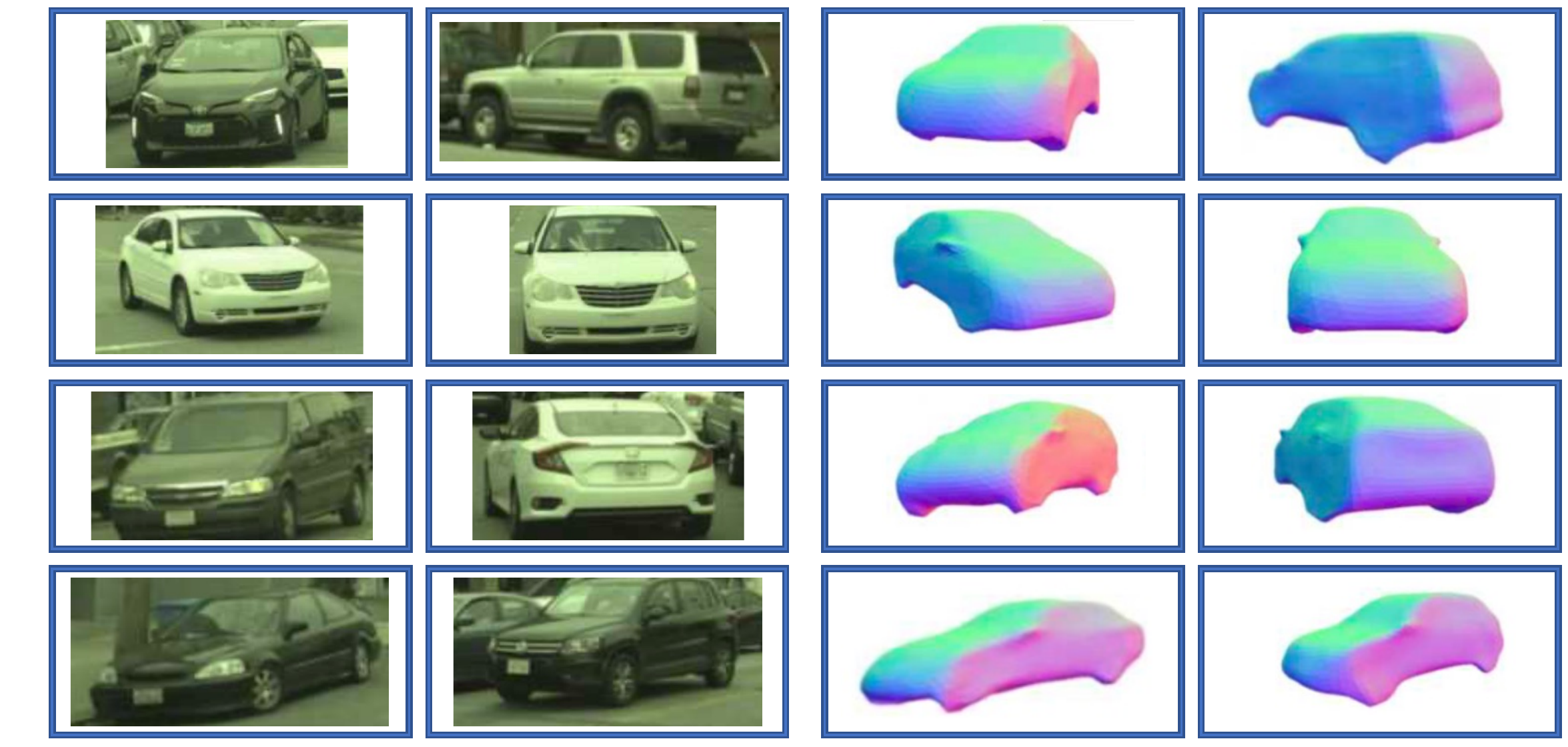}
\label{fig:assets}
\end{minipage}
\end{center}
\cutcaptionup
\vspace{-.5em}
\caption{\textbf{Realistic 3D \yun{assets creation.}} Left: multi-view multi-sensor reconstruction network; Right: 3D asset samples. For each sample we show one of the source images and the 3D mesh.}
\label{fig:architecture_assets}
\vspace{-.5em}
\cutcaptiondown
\end{figure*}

\paragraph{Simulation for Robot Learning:}
Sensor simulation has received wide attention in the literature~\cite{ros2016synthia,dosovitskiy_carla_2017,richter_playing_2016,alhaija_augmented_2017,savva2017minos,savva2019habitat,bousmalis2018using,coumans2016pybullet,xia_gibson_2018,kolve2017ai2,savva2019habitat,li2019putting} for its applications in training and evaluating robotic agents. 
Sensor simulation systems should be efficient and scalable in order to enable such applications. 
 Many automatic approaches \cite{savva2017minos,savva2019habitat,coumans2016pybullet,xia_gibson_2018,kolve2017ai2} have been proposed  to generate  indoor environments.
Unconstrained outdoor scenes such as the urban driving setting tackled here bring additional challenges due to the scale of the scene, weather, lighting, presence of fast moving objects, and large viewpoint changes arising from sensor motion.
In the context of autonomous driving, simulation engines \cite{dosovitskiy_carla_2017,ros2016synthia,richter_playing_2016} based on rendered 3D models allow the combinatorial generation of scenarios with varying configurations of vehicle attributes, traffic, and weather conditions.
However, these methods often have limited diversity in scene content due to the manual design of 3D assets and still have a Real2Sim gap.
Data-driven sensor simulation offers a scalable alternative that can capture the complexity of the real world.
Many methods \cite{li_aads_2019,fang_augmented_2019,Tallavajhula-2018-106471,yang_surfelgan_2020,amini_learning_2020,siva2020lidarsim} have been proposed to directly leverage real-world data for sensor simulation in the autonomous driving domain, 
typically by augmenting existing recorded data to generate corresponding sensor measurements for novel scene configurations. 
However, previous works either focus on LiDAR \cite{siva2020lidarsim,fang_augmented_2019,Tallavajhula-2018-106471}, rely on CAD model registration, constraining the set of dynamic objects that can be simulated
\cite{li_aads_2019},
or require additional effort to scale to high-resolution images \cite{yang_surfelgan_2020}.
In contrast, we combine data-driven simulation techniques
with the image-based rendering techniques in simulation engines.
This enables us to construct a scalable, geometrically consistent, and 
\sm{realistic}
camera simulation system.

\cutparagraphup
\paragraph{Image Synthesis and Manipulation:}
Image synthesis and manipulation methods offer another route to sensor simulation. %
Existing work mainly focused on generating 2D images from intermediate representations including scene graphs~\cite{johnson2018image,hong2018inferring},
surface normal maps~\cite{wang2016generative},
semantic segmentations~\cite{isola2017image,zhu2017unpaired,chen_photographic_2017,wang_high-resolution_2017,qi_semi-parametric_2018,park2019semantic,mo2018instagan},
and images with different styles~\cite{karras2019style}.
These methods create high-resolution images but with noticeable artifacts in texture and object shape.
Rather than generating the full image in one shot, \cite{lin_st-gan_2018,hong_learning_2018,lee2018context} utilize a conditional image generator for scene manipulation.
In particular, \cite{lin_st-gan_2018} proposed a spatial-transformer GAN  that overlays the target objects on top of  existing scene layouts by iteratively adjusting  2D affine transformations.
\cite{hong_learning_2018} introduced a hierarchical image generation pipeline that is capable of inserting and removing one object at a time.
{This improves realism, but using purely a network-based image synthesis approach has difficulty handling complex physics such as lighting changes. 
\cite{bhattad2020cut} attempts to combine data driven approaches with graphics knowledge, using an image-based neural renderer and image decomposition to  improve the synthetic result.}
{Our work builds on this direction of leveraging graphics with real world data.}
{We perform image-based rendering and neural in-painting to adjust for differences between the
original image and the image texture of the inserted actor.}
Furthermore, GeoSim is \sm{3D-layout-aware, allowing for controllable and realistic scene modification.}

\cutparagraphup
\paragraph{\shivam{Video Synthesis and Manipulation:}}
\sm{Image simulation alone is insufficient for generating new scenarios with realistic video.
One way prior \shivam{works} \shivam{have} extended image synthesis approaches to video generation is by including the past and adding regularization to ensure temporal consistency for realistic object \yun{motion}.
Conditional video generation methods  \cite{wang2018vid2vid,mallya2020world, chan2019everybody, gafni2019vid2game} take 
\shivam{sequential} semantic masks, \shivam{depth maps or trajectory pose data as inputs}, which can then be semantically modified to generate the current video frame.
\cite{ehrhardt2020relate} performs 2D-aware image composition via generative modeling of objects and learned dynamics.
Automatic video manipulation approaches \cite{lee2019inserting, ibrahim2020inserting} \shivam{insert foreground objects or object videos into existing videos} in a seamless manner.  
Unlike most prior work, our image-composition approach is {3D-layout-aware} and handles occlusions. Thus, by combining our image composition with automatic trajectory generation methods \cite{schulz_interaction-aware_2018,suo2021trafficsim}, we easily extend to automatic and scalable controllable video simulation with high realism.}

\cutparagraphup
\paragraph{3D Reconstruction and View Synthesis:}

Our neural network-based 3D asset creation step reconstructs 3D shape from camera imagery and LiDAR in order to synthesize novel views of dynamic objects.
View synthesis and 3D reconstruction are well-studied open problems  \cite{tewari_state_2020}, with varying approaches on the relationship between geometry and appearance and possible geometric representations.
Image-based rendering methods~\cite{DeepBlending2018} focus on combining 2D textures to directly render novel views. Appearance flow-based approaches~\cite{zhou_view_2016,park_transformation-grounded_2017,ferrari_multi-view_2018} seek to learn unconstrained pixel-level displacements, whereas~\cite{chen_monocular_2019,zhu_visual_2018} encode geometric information in latent representations and~\cite{kanazawa_learning_2018} takes advantage of strong shape priors.
Recently, advancements in differentiable rendering~\cite{liu_soft_2019,kato_neural_2017} and open-source libraries have enabled classical graphics rendering to serve as an optimizable module, allowing for better learning of 3D and visual representations~\cite{chen2019dibrender, zhang2020image}.

\cutsectionup
\section{Realistic 3D Assets Creation}
\cutsectiondown
\label{sec:assets}

In this paper we propose a novel image manipulation approach that inserts dynamic objects into an existing video footage and generates a 
\sm{high-quality}
video of the augmented scene that is geometrically and semantically consistent with the scene.
Key to the success of such \yun{an} approach is the availability of realistic 3D assets that contain accurate pose, shape and texture.
Here we argue that rather than using artists to create these assets, we can leverage data captured by self-driving vehicles to reconstruct the objects around us. This is a much more scalable approach, as many self-driving datasets are available \cite{argoverse,waymo,kitti}%
, each containing many thousands of unique assets that could potentially be reconstructed.
{
In Sec.~\ref{sec:multi_sensor_assets} we first describe how we leverage both LiDAR and camera sensor data from multiple viewpoints to generate realistic 3D vehicle assets using an asset reconstruction network. 
Sec.~\ref{sec:assets_learning} describes our self-supervised learning procedure to train the network.}

\cutsubsectionup
\subsection{Multi-Sensor 3D Asset Reconstruction}
\cutsubsectiondown
\label{sec:multi_sensor_assets}

Reconstructing 3D dynamic objects in the wild %
is 
challenging:
dynamic objects are often partially observed due to the sparsity of the sensor observations 
and 
occlusions, they are seen from a 
{limited}
set of viewpoints, and they appear distorted due to lighting and shadows.
To tackle these challenges, we propose a novel, learning-based, multi-view, multi-sensor reconstruction approach for 3D dynamic objects that does not require any ground-truth 3D-shape for training.
Instead, we exploit weak annotations in the form of 3D bounding boxes, which  %
 are readily available in most self-driving datasets.

\begin{figure*}[t]
\centering
\includegraphics[width=\textwidth,trim={0 8cm 0 5cm},clip]{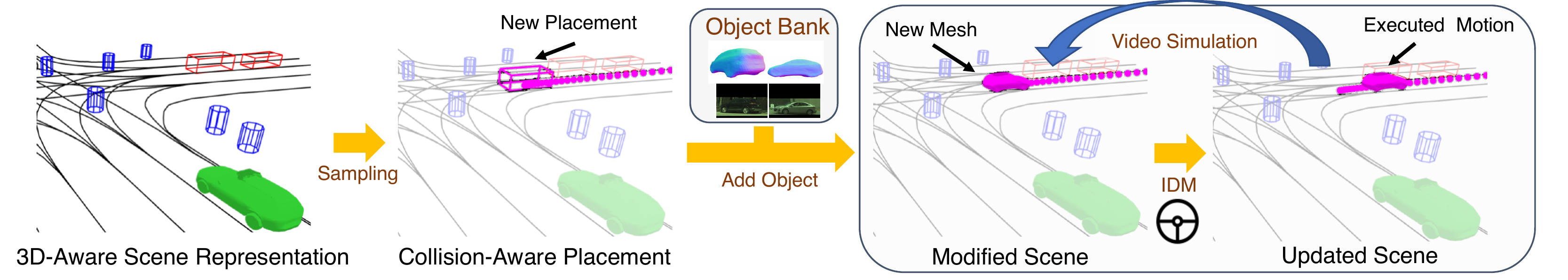} \\
\cutcaptionup
\caption{
{\textbf{3D-aware object placement, segment retrieval, and temporal simulation.}} }
\cutcaptiondown
\label{fig:fig_placement}
\vspace{-.5em}
\end{figure*}

More formally, let $\{\mathbf{B}_{i,j}\}_{\forall j}$ be the set of 3D bounding boxes where the i-th object is visible over  $j$ views in the recorded snippet.
Let $\{\mathbf{I}_{i,j}\}_{\forall j}$ be the {associated} set of {cropped} images, {and } 
$\{\mathbf{X}_{i,j}\}_{\forall j}$ be the associated set of lidar points recorded inside $\{\mathbf{B}_{i,j}\}$,  transformed to a single canonical frame and
let $\mathbf{X}_{i}$  be the set of {aggregated} LiDAR points across all views.
Our 3D reconstruction network then processes the LiDAR points and image inputs in two streams that are later fused to produce the 
shape 
of the object. We refer the reader to Fig.~\ref{fig:architecture_assets} for an illustration.
{We represent the shape as} a 3D mesh $\mathbf{M}_i = \{ \mathbf{V}_i, \mathbf{F}_i \}$ where $\mathbf{V}_i$ and $\mathbf{F}_i$ are the faces and vertices of the mesh, respectively. 
In addition, we also store $\{\mathbf{I}_{i,j}, \mathbf S_{i,j}\}_{\forall j}$ to encode object appearance, where $\mathbf S_{i,j}$ is the extracted object's silhouette obtained from a pre-trained instance segmentation model~\cite{kirillov2020pointrend}.
We use this later on to perform novel-view warping.

\cutparagraphup
\paragraph{Network Architecture:}
Our backbone consists of two submodules.
A convolutional network takes each cropped camera image as input and outputs a corresponding feature map. 

The feature maps from multiple cameras %
are then aggregated into a one-dimensional latent representation using max-pooling.
A similar latent representation is extracted from the input  LiDAR point cloud using a PointNet network \cite{qi2016pointnet}.
The LiDAR and camera features are then passed   through an MLP to output the final shape.
\yun{Instead of employing a learned PCA shape space from CAD models to predict the shape of cars \cite{KunduLR18},}
we \yun{take inspiration from \cite{kanazawa_learning_2018} and} parameterize the 3D shape as a category-specific \sm{canonical} mean shape \sm{with per-vertex deformations.} 
{This parameterization} encodes a categorical prior and ensures the completeness of the shape under partial observations.

\cutsubsectionup
\subsection{Self-Supervised Learning}
\cutsubsectiondown
\label{sec:assets_learning}
Note that we do not have supervision about the shape.
We thus train our approach end-to-end in an 
{self-supervised}
manner to obtain the parameters of the reconstruction network and the mean shape. %
Our training objective encodes the agreement between the 3D shape and the camera and LiDAR observations,  as well as a regularization term. %
\begin{align*}
	\ell_{\textrm{total}} =  \sum_i \{ \ell_{\textrm{sil}}(\mathbf{M}_i; \mathbf{P}_i, \mathbf{I}_i) \;+\; \ell_{\textrm{lidar}}(\mathbf{M}_i; \mathbf{X}_i) \;+\;\ell_{\textrm{reg}} (\mathbf{M}_i) \}
\end{align*}
where $i$ {ranges over all the training objects. }
The \emph{silhouette loss} measures the consistency between the  2D silhouette (automatically generated using the state-of-the-art object segmentation method PointRend \cite{kirillov2020pointrend}) and the silhouette of the rendered 3D shape.
\[\ell_{\textrm{sil}}(\mathbf{M}_i; \mathbf{P}_i, \mathbf{I}_i) = \sum_j\norm{\mathbf{S}_{i,j} - \tau(\mathbf{M_{i,j}}, \mathbf{P_{i,j}}) }_{2}^2\]
where $\mathbf{S_{i,j}}  \in \R^{ H \times W}$ is the 2D silhouette{,  $j$ ranges over multiple views}, and  $\tau(\mathbf{M}, \mathbf{P})$ is a differentiable neural rendering operator \cite{chen2019dibrender} that renders a differentiable mask on the camera image given a projection matrix $\mathbf{P}$. %
The \emph{LiDAR loss} represents the consistency between the {accumulated} LiDAR point cloud and the mesh vertices, defined as the asymmetric Chamfer distance
\[\ell_{\textrm{lidar}}(\mathbf{M_i}, \mathbf{X_i}) = \sum_{\mathbf{x} \in \mathbf{X}_i} \min_{\mathbf{v} \in \mathbf{V}_i} \norm{\mathbf{x} - \mathbf{v}}_{2}^2\]
In addition, we also minimize a set of \emph{regularizers} to enforce prior knowledge over the 
{predicted}
3D shape, namely local smoothness on the vertices as well as normals. 
This includes 
1) a Laplacian regularization which preserves local geometry and prevents intersecting faces;
2) mesh normal regularization which enforces smoothness of local surfaces;
3) edge regularization which penalizes long edges. 
Please refer to \shivam{supp.} material for details.

\cutsectionup
\section{Geometry-Aware Image Simulation}
\cutsectiondown
\label{sec:sim}
Here we describe our image simulation by composition approach that places novel objects into an existing 3D scene and generates a 
\sm{high-quality}
video sequence of the composition.
Our approach takes as input camera video footage, LiDAR point clouds, and an HD map in the form of a lane graph
and automatically outputs a 
video with novel objects
inserted into the scene.
Note that the input sensory data and HD maps we employ are readily available in most self-driving platforms, which are the application domain we tackle in this paper.
Importantly, our simulation takes into account  both geometric occlusions by other actors and the background, plausibility of the locations and motions as well as the interactions  with other dynamic agents and thus avoids collision for the newly inserted objects.

Towards this goal, we first infer the location of all objects in the  scene by performing 3D object detection and tracking \cite{liang2020pnpnet}. %
For each new object to be inserted we select where to place it  as well as which asset to use based on the HD map and the existing detected traffic.
We then utilize an intelligent traffic model for our newly placed object such that its motion is realistic, takes into account the interactions with other actors and avoids collision. The output of this process defines the new scenario to be rendered.
We then use a novel-view rendering with 3D occlusion reasoning w.r.t.~all elements in the scene, to create the appearance of the novel objects in the new image.
Finally, we utilize a neural network to fill in the boundary of the inserted objects, create any missing texture and handle inconsistent lighting. %
Fig.~\ref{figure:pipeline} outlines our approach.

\begin{figure*}[t]
\centering
\includegraphics[width=\textwidth]{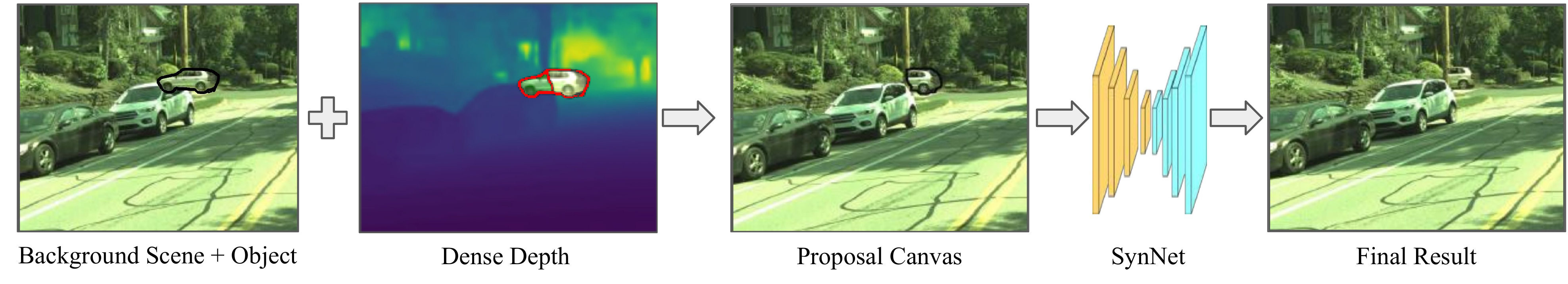} \\
\cutcaptionup
\vspace{-.5em}
\caption{\textbf{Geometry-aware composition with occlusion reasoning followed by an image synthesis module.}}
\label{fig:occlusion_inpainting}
\vspace{-.5em}
\cutcaptiondown
\end{figure*}

\cutsubsectionup
\subsection{Scenario Generation}
\cutsubsectiondown
 \label{sec:placement}

We want to place  new objects in existing images such that they are plausible in terms of their scale, location, orientation and motion.
Towards this goal, we design a 3D sampling procedure, which takes advantage of the priors we have about how vehicles behave in our cities.  Note that 
\sm{it is difficult to achieve similar levels of realism with 2D object insertion.}
We thus exploit HD maps that contain the location of the lanes in bird's eye view (BEV),
and  parameterize the object placement as a tuple $(x,y,\theta)$ defining the object center and orientation in BEV, which we later convert to a 6DoF pose using the local ground elevation.

Note that our object samples should have
realistic physical interactions with existing objects, respect the flow of traffic, and be visible in the camera's field of view.
To achieve this, we randomly sample a placement $(x,y)$ from the lane regions lying within the camera's field of view and retrieve the orientation from the lane. We reject all samples that result in collision with other actors or background objects.
The aforementioned process provides the placement of the object in the initial frame.
To simulate plausible placements over time for video simulation,
we use the Intelligent Driver Model (IDM) \sm{\cite{schulz_interaction-aware_2018, suo2021trafficsim}}
fitted to a kinematic model following \cite{Gonzales2016AutonomousDW},
to update the simulated object's state for realistic interactions with surrounding traffic.
Fig.~\ref{fig:fig_placement} depicts the full procedure of placement and kinematics simulation.

So far we have selected where to place an object and how is going to move, but we still need to select which object to place. In order to minimize the artifacts when rendering our assets, we propose to  retrieve objects from the asset bank that were viewed with  similar viewpoints and distance to the camera in the original footage.  The former avoids the need to deal with large unseen object regions while the latter avoids utilizing assets that have been captured at lower resolution. %
Please refer to the 
\shivam{supp.}
 for the specific scoring criteria.
Objects are then sampled (as opposed to a hard max) according to a categorical distribution weighted by their inverse score.
Once a segment is retrieved for a desired placement, we perform collision checking  using the retrieved object shape to ensure that the placement is  valid. %

\cutsubsectionup
\subsection{Occlusion-Aware Neural Rendering}
\cutsubsectiondown
\label{sec:warping}

Now that we have selected a source object and its corresponding camera image based on the segment retrieval mechanism defined above, we proceed to render this source object into the target scene. Since the object's target pose might differ from the original observed poses,  we cannot simply paste the image segment from the source to the  target. Thus we proposed to utilize the asset's 3D shape   to warp the source to the novel target view.

\cutparagraphup
\paragraph{Novel-view Warping:} Let $\mathbf{M}$ be the selected object's 3D mesh, $\mathbf{I}_{\textrm{s}}$ be the source object's camera image,
and $\mathbf{P_{\textrm{s}}}/ \mathbf{P_{\textrm{t}}}$ be the source/ target camera matrices.
We first render $\mathbf{M}_{\textrm{s}}$ at the selected target viewpoint to generate the corresponding target depth map, $\mathbf{D_{\textrm{t}}}$. Then using the rendered depth map and source camera image $\mathbf{I}_{\textrm{s}}$, we generate the object's 2D texture map using the inverse warping operation \cite{lsiTulsiani18,jaderberg_spatial_2016} as:
\begin{align*}
	\mathbf{I_{\textrm{t}}} &= \mathbf{I_{\textrm{s}}}(\pi(\pi^{-1}(\mathbf{D_{\textrm{t}}}, \mathbf{P_{\textrm{t}}}), \mathbf{P}_{\textrm{s}})) \text{\  , where \ \ \ } 	\mathbf{D_{\textrm{t}}} = \psi(\mathbf{M}, \mathbf{P_{\textrm{t}}}),
\end{align*}
 $\psi$ is a differentiable neural renderer~\cite{chen2019dibrender} that produces a depth image given the 3D mesh $\mathbf{M}$ and camera matrix $\mathbf{P}$; $\pi$ is the perspective projection and $\pi^{-1}$ is the inverse projection that takes the depth image and camera matrix as input and outputs the 3D points.

\cutparagraphup
\paragraph{Shadow Generation:}
Inserting an object into a scene will not \sm{only change}
the pixels where the object is present, but can also change the rest of the scene (i.e. shadows and ambient occlusion). 
We improve the perceptual quality of the image by approximating these effects with image based rendering.
\yun{While recent works~\cite{wang_people_2020,zhang2019shadowgan} learn shadow synthesis from scene context with a neural network, we render shadows with a graphics engine as geometry is available.} 
To estimate the shadow casted by each inserted object, we construct a virtual scene with the inserted object and a ground plane and exploit image-based rendering~\cite{debevec2008rendering}, where the environment light comes from a real-world HDRI. 
We render the scene with and without the inserted objects, and add the shadow by blending in the background image intensities with the ratio of the two rendered images intensities. 
\yun{As lighting estimation by manually waving a shadow-casting stick~\cite{chuang2003shadow} is not applicable,} we select a cloudy HDRI to cast shadows.
In practice, we find this produces reasonable results.
\yun{Please refer to the \shivam{supp.} for illustration. }

\cutparagraphup
\paragraph{Occlusion Reasoning:}
An inserted object must respect occlusions from the existing scene elements.
Vegetation, fences, and other dynamic objects, for example, may have irregular or thin boundaries,
complicating occlusion reasoning.
We employ a simple strategy to determine occlusions of the inserted objects and their shadow in the target scene by
comparing their depth w.r.t the depth map of the existing 3D scene
\yun{ (see Fig.~\ref{fig:occlusion_inpainting})}.
To achieve this, we first estimate the target image's dense depth map through a depth completion network~\cite{Chen_2019_ICCV}.
The input is the RGB image and a sparse depth map acquired by projecting the LiDAR sweep onto the image.
Using the rendered depth of the object,
the occlusion mask is then computed by evaluating for  each object pixel if the target image's depth is smaller than the corresponding object  pixel's depth.

\cutparagraphup
\paragraph{Post-Composition Synthesis:}
After occlusion reasoning, the rendered image may still look unrealistic as the inserted segment may have inconsistent illumination and color-balancing w.r.t the target scene, discrepancies at the boundaries, and missing regions that were not available in the source view.
To solve these issues, we leverage an image synthesis network %
to naturally blend the source segment to the target scene (see Fig.~\ref{fig:occlusion_inpainting}).  Our network takes the target background image $\mathbf{B}_t$, \yun{rendered target object}  $\mathbf{I}_t$ as well as the object binary silhouette $\mathbf{S}_t$ as input, and outputs the final image that naturally composites the background and rendered object.
Our synthesis network architecture is similar to \cite{yu_free-form_2019}, which is a generative image in-painting network %
except that we take the rendered object mask as additional input. %
Our network is trained using images with instance segmentation masks inferred by \cite{kirillov2020pointrend} in the target scene.
Data augmentation, including random occlusion, color jittering, 
random contrast and saturation is applied to mimic the differences among real-world images.
Two loss functions are adopted, namely perceptual loss \cite{DBLP:journals/corr/JohnsonAL16} to ensure the generated output's fidelity, as well as GAN loss to boost the realism of the in-painted region as well as the lighting consistency. Please refer to 
\shivam{supp.}
 for more details.

\begin{figure*}[t]
    \centering
    \includegraphics[width=1\textwidth]{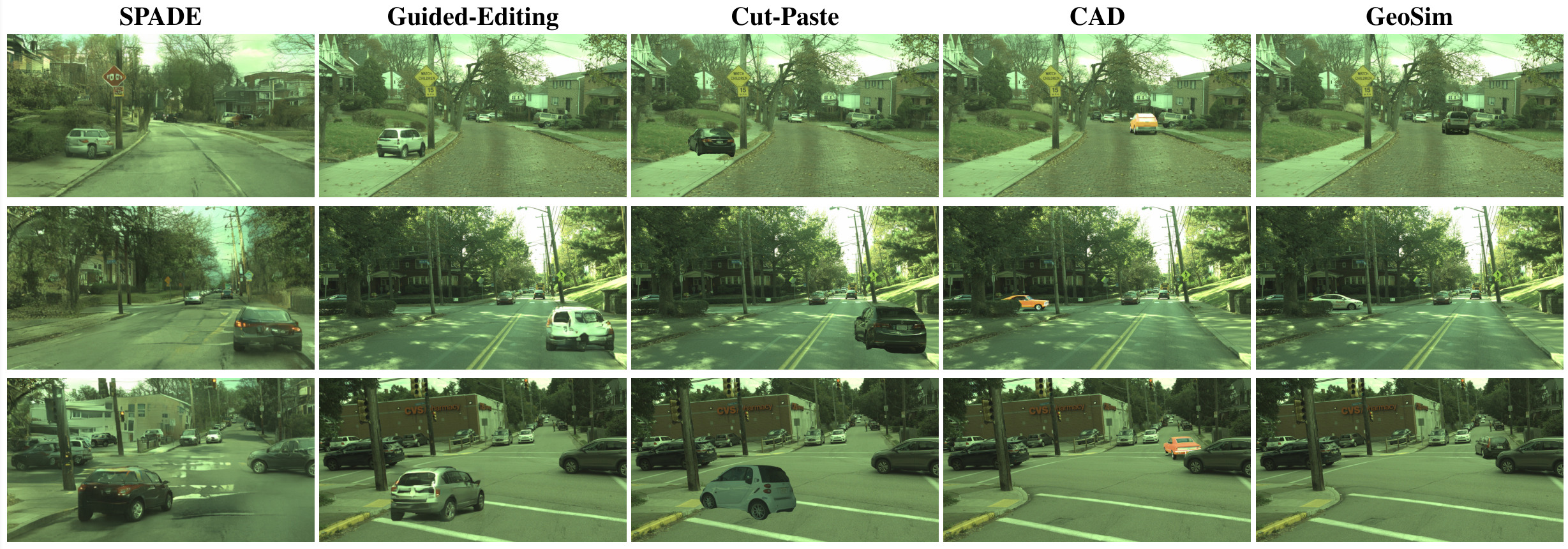} \\
    \vspace{-.25em}
    \caption{\textbf{Qualitative comparison of image simulation approaches.}}
    \cutcaptiondown
    \vspace{-.25em}
    \label{fig:qual-baselines}
    
    \end{figure*}

\cutsectionup
\section{Experimental Evaluation}
\cutsectiondown
In this section we first introduce our experimental setting.
We then compare GeoSim against a comprehensive set of image simulation baselines 
in visual realism through perceptual quality scores and human A/B tests, and in downstream tasks such as semantic segmentation.
We also showcase our method generating multi-camera and temporally consistent video simulation.
While our method can be adapted to handle most rigid objects, in our experiments we showcase vehicles, the most 
relevant objects in self driving.

\cutsubsectionup
\subsection{Experimental Details}
\cutsubsectiondown
\label{sec:exp_setting}

{We utilize two large-scale self-driving datasets (UrbanData  and Argoverse \cite{argoverse}) to showcase GeoSim.

\cutparagraphup
\paragraph{UrbanData:} We collected a real-world dataset by having a fleet of self-driving cars drive in two major cities
in North America. %
We labeled  
16,500
 snippets, where each snippet contains 25 seconds (\textasciitilde{}250 frames, sampled at 10Hz) of video  with 7 cameras, a 64-beam LiDAR sensor, and HD maps. 
We 
use 4500 for reconstruction and synthesis network training, 7000 for depth completion training, and 5000 for perceptual quality and downstream evaluation. Please see supp. for the full breakdown.

\cutparagraphup
\paragraph{Argoverse:} {We also evaluate on the
Argoverse training split which
contains 65 snippets from 2 different cities. We use the provided HD maps for vehicle placement.
We directly adopt the 3D assets built from UrbanData, as Argoverse is too small for diverse asset creation.  We train our image synthesis network on Argoverse, where \sm{80k} frames \yun{are} sampled for  training and \sm{16k}
are \yun{sampled} for evaluation.

\cutparagraphup
\paragraph{Asset Bank Creation:} We created automatically a large object bank of  \textasciitilde{}   8000 vehicles, from cameras, LiDAR data and 3D bounding boxes using our 3D reconstruction network on UrbanData.
Each successfully reconstructed object is registered in our 3D asset bank, with its 1) 3D mesh; 2) images; and 3) object pose in ego-vehicle-centric coordinates.
We use a pre-trained instance segmentation to get the inferred instance mask~\cite{kirillov2020pointrend}, a LiDAR detector \cite{liang2020pnpnet} to acquire other actors' bounding boxes for collision avoidance (Sec. \ref{sec:placement}), and a depth completion network~\cite{Chen_2019_ICCV} to get dense depth for occlusion reasoning (Sec. \ref{sec:warping}). %

\cutparagraphup
\paragraph{Baselines:}
\label{sec:exp_baselines}
We compare our method against several deep learning based end-to-end 2D image synthesis and augmentation baselines~\cite{lee2018context,wang_high-resolution_2017,hong_learning_2018, park2019semantic}.
Unlike GeoSim, these methods cannot perform placement directly and require an input mask based on the \yun{object's} shape and pose that \yun{denotes} the area to synthesize.
We therefore use \cite{lee2018context} to insert object instances at the semantic level in a background semantic image.
We then generate high resolution images from this augmented scene representation with three different approaches:
(1) Holistic image generation (\textbf{"SPADE''}): we use the state-of-the-art conditional image generation model SPADE~\cite{park2019semantic} to generate the entire image  given the semantic mask. 
(2) Retrieval-based generation (\textbf{"Cut-Paste''}): given the new object's 2D mask, we retrieve the most similar example from a bank of 2D object images. The similarity is defined using semantic mask IOU. The rest of the background comes from the corresponding real image;
and (3) Guided semantic image editing (\textbf{"Guided-Editing''}): we use  \cite{hong_learning_2018} to in-paint the tight bounding box region of the
added object.
Additionally, we compare  against a graphics-based CAD model insertion baseline (\textbf{"CAD''}), in spirit of  \cite{alhaija2018augmented} with the following differences: 1) we use our 3D placement in order to produce more realistic layout-aware insertion; 
2) unlike the original work, we do not have environment lighting \sm{maps}
and \sm{instead} 
use a HDRI captured on a cloudy day.
}

\cutsubsectionup
\subsection{Perceptual Quality Evaluation}
\cutsubsectionup
\label{sec:exp_realism}

\begin{figure}[t]
    \centering
    \includegraphics[width=0.5\textwidth]{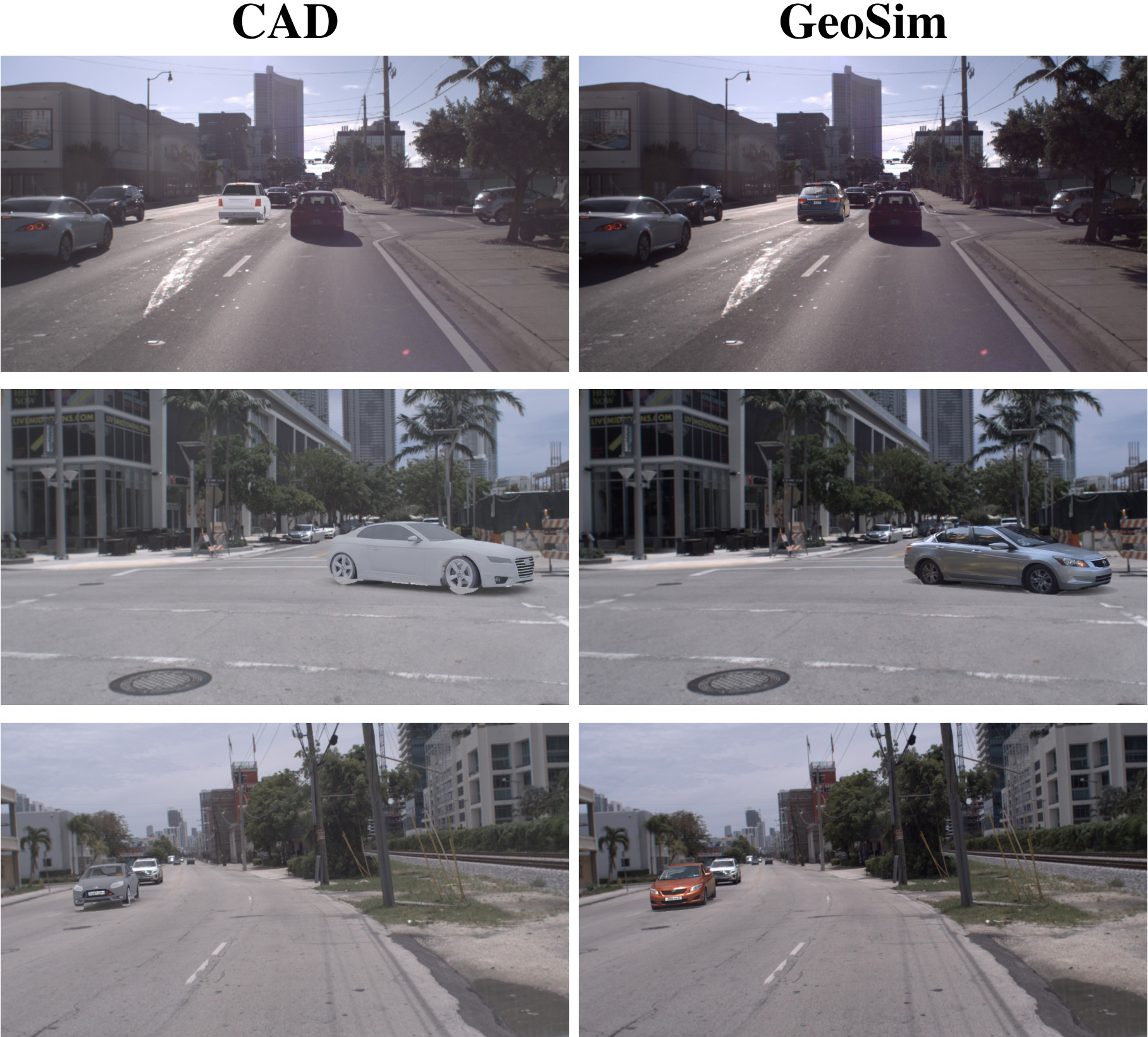} \\
    \vspace{-.25em}
    \caption{\textbf{Qualitative comparison of image simulation {approaches} \shivam{on Argoverse dataset}.}}
    \cutcaptiondown
    \vspace{-.5em}
    \label{fig:qual-baselines-argo}
    
    \end{figure}

\paragraph{Human Study:}  To verify the realism of our approach, we conduct a human A/B test, where  we show a pair of images generated from different approaches on the same background image, one from GeoSim and another one from a competing algorithm. We then ask the human judges to click the one they believe is more realistic.
In total, 13 human judges participated and labeled \textasciitilde{} 1500 image pairs. %
Tab.~\ref{tab:baseline} shows the human preference score for each algorithm, which measures the percentage of participants who prefer our GeoSim results over each baseline method.
Results on Argoverse are presented in Tab.~\ref{tab:argo}.
The A/B test confirms that our method produces drastically more realistic images than the baselines. 
\yun{The minimum $p$-value in the A/B test is 1.64e-18, demonstrating statistical significance.} %
Please see the detailed A/B test interface and instructions 
in supp. %

\cutparagraphup
\paragraph{Perceptual Quality Score:} We further use the Fr\'echet Inception Distance (FID)~\cite{FID} between the synthesized images and the ground-truth images  as  an automatic measure of image quality.
We report the FID on the full image
{for GeoSim and the baselines} in Tab.~\ref{tab:baseline}.
Our method 
significantly outperforms all 
competing methods on
FID. 

\cutparagraphup
\paragraph{Qualitative Comparison:}
Fig.~\ref{fig:qual-baselines} 
\shivam{compares} simulated images.
Note that GeoSim is significantly more realistic than the baselines. 
While one can easily and quickly detect the added object in other methods due to unrealistic generation \shivam{with} smeared cars ({"SPADE''}, {"Guided-Editing''}), or geometrically invalid results ({"Cut-Paste''}), or unrealistic appearance ({"CAD''}), one must look closely at {GeoSim} 
images to distinguish  the added objects from the real ones. 
\shivam{In Fig.~\ref{fig:qual-baselines-argo}, we show qualitative examples on Argoverse, where GeoSim obtains very high visual quality.}
This demonstrates GeoSim's 
potential to 
generalize across datasets.

\cutparagraphup
\paragraph{Effect of Rendering Approach:}
We evaluate the importance 
of using a hybrid rendering module proposed in our method, compared to using solely physics-based rendering or 2D synthesis %
with 3D placement constant across all approaches).
As shown in Tab.~\ref{tab:ablation}, 
our proposed \yun{geometry-aware} synthesis significantly outperforms all other approaches on human scores.
Additionally, enhancing hybrid-rendering with shadows
significantly boosts the realism for humans, but such improvements are not reflected in FID score. This suggests there still exists a gap between computational perceptual quality measurements and humans' criteria. 
\sm{Please see supp. for ablation of other GeoSim components.}

\cutparagraphup
\paragraph{Video Simulation:}
We 
showcase in the supp. video GeoSim's ability to
simulate \sm{highly realistic and}
temporally consistent video 
for multiple cameras.

\cutparagraphup
\paragraph{Failure Cases:}
\yun{
While most GeoSim-simulated images 
\sm{are high-quality,
there is room for improvement.}
 We find four major failure cases:
(1) incorrect occlusion relationships in a complex scene,
(2) irregular reconstructed mesh,
(3) inaccurate object pose, usually caused by map error 
and (4) illumination failure \shivam{due to} 
illumination differences between rendered segment and target scene.
Besides, we also notice blank pixel artifacts in long range video simulation, which 
\shivam{are caused by inverse warping textures
from source viewpoints which are \yun{far} from the target viewpoint.}
Please refer to the supp. for qualitative examples.
}

\begin{table}[]
    \centering
    \begin{tabular}{ccc}
    \hline
    \specialrule{.2em}{.1em}{.1em}
    Method         & Human Score \shivam{(\%)} & FID   \\ \hline
    SPADE~\cite{park2019semantic}         & 99.3       & 43.2 \\
    Guided Editing~\cite{hong_learning_2018} & 94.3       & 20.3 \\
    Cut-Paste~\cite{dwibedi2017cut}      &  98.5   & 22.1 \\
    CAD~\cite{alhaija_augmented_2017}  &  94.3    & 17.3 \\
    GeoSim         &  \textbf{-}    & \textbf{14.3} \\\hline
    \end{tabular}
    \vspace{-0.25em}
    \cuttablecaptionup
    \caption{\textbf{Perceptual quality evaluation.}  Human score: 
    \shivam{\%} 
    of participants who \shivam{prefer} our  GeoSim results over baseline.}
    \cuttablecaptiondown

    \label{tab:baseline}
    \end{table}
    
\begin{table}[]
    \scalebox{1}{
    \centering
    \begin{tabular}{ccccc}
    \hline
\specialrule{.2em}{.1em}{.1em}
    Approach &   Shadow & Human Score \shivam{(\%)} & FID         \\ \hline
    Physics              & Yes &         94.2     & 17.3 \\ 
    2D Synthesis    &  - &         75.7     & \textbf{13.7}  \\ 
       Geo Synthesis &  No &       71.9     & \textbf{13.7} \\ 
    Geo Synthesis & Yes &     \textbf{-}    & 14.3  \\ \hline
    \end{tabular}}
    \label{tab:ablation_table}
   \cuttablecaptionup
   \vspace{-0.25em}
    \caption{\textbf{Ablation on rendering options for GeoSim.} Human score: 
    \shivam{\%} 
    of participants who \shivam{prefer} our  GeoSim results over baseline.
    \label{tab:ablation}}
	\cuthalftablecaptiondown
    \vspace{-.25em}

\end{table}

\begin{table}[]
\centering
    \begin{tabular}{ccc}
    \hline
\specialrule{.2em}{.1em}{.1em}
    Method  & Human Score  \yun{(\%)}   &  FID  \\ \hline
    CAD    & 84.0 &   28.3         \\ 
    GeoSim  & -&    \textbf{24.5}    \\ \hline
    \end{tabular}
    \vspace{-.25em}
	\cuttablecaptionup
    \caption{\textbf{Results on Argoverse}. Human score: \% of participants who prefer our GeoSim results over baseline.
    }
    \cuttablecaptiondown
    \label{tab:argo}
    \end{table}

\begin{table}[]
    \centering
        \begin{tabular}{ccccc}
        \hline
    \specialrule{.2em}{.1em}{.1em}
    Method & \multicolumn{2}{c}{PSPNet~\cite{zhao2017pspnet}} &  \multicolumn{2}{c}{DeepLabv3~\cite{chen2017rethinking}}  \\
                    & mIOU & carIOU & mIOU & carIOU \\ \hline
        Real        & 93.5       & 87.8          & 94.0          & 88.7     \\         
        Real+GeoSim & \textbf{95.3}        & \textbf{91.2}          & \textbf{94.2}          & \textbf{89.2}       \\       \hline
        \end{tabular}
        \cuttablecaptionup
        \vspace{-.25em}
        \caption{\textbf{Sim2Real on semantic segmentation.} }
        \cuthalftablecaptiondown
        \label{tab:sim2real}
    \end{table}

    \begin{figure}[t]
    \centering
    \setlength{\tabcolsep}{1pt}

    \begin{tabular}{ccc}
        \textbf{{Background}} & \textbf{Augmented}  & \textbf{Augmented Label}\\
        \includegraphics[width=0.33\linewidth,trim={7cm 1cm 8cm  1cm},clip]{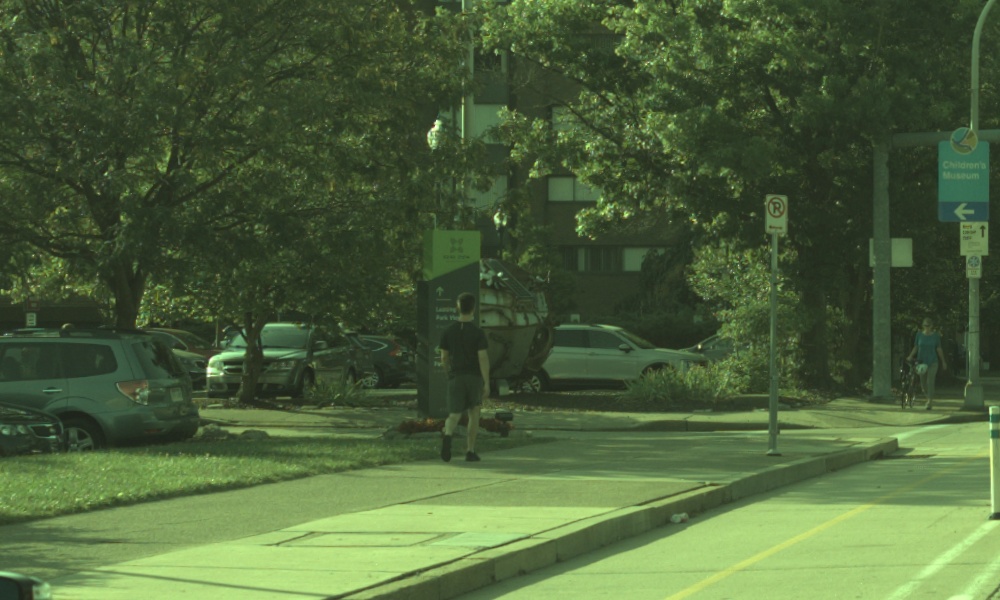}
        & \includegraphics[width=0.33\linewidth,trim={7cm 1cm 8cm  1cm},clip]{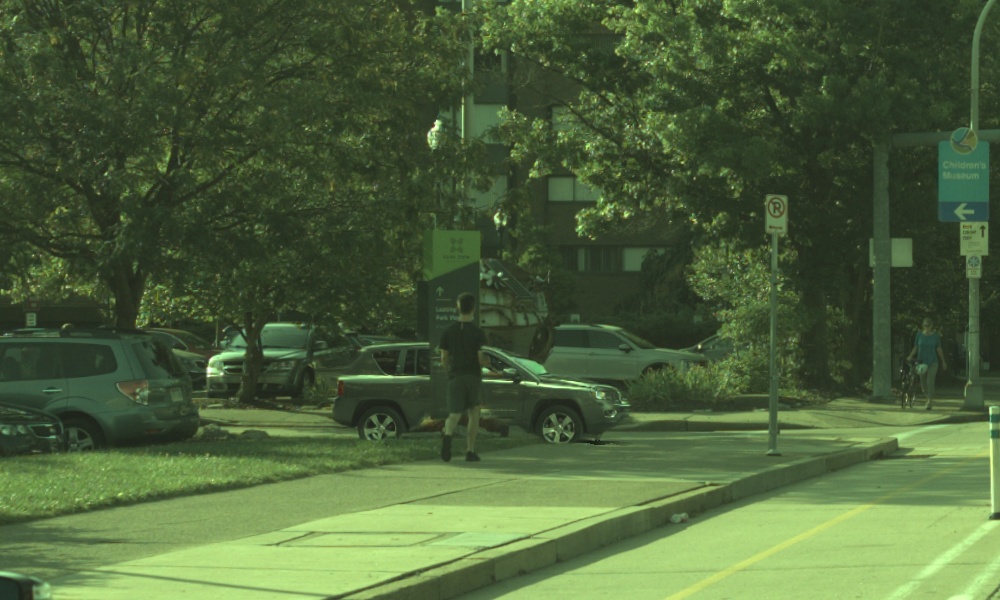}
        & \includegraphics[width=0.33\linewidth,trim={7cm 1cm 8cm  1cm},clip]{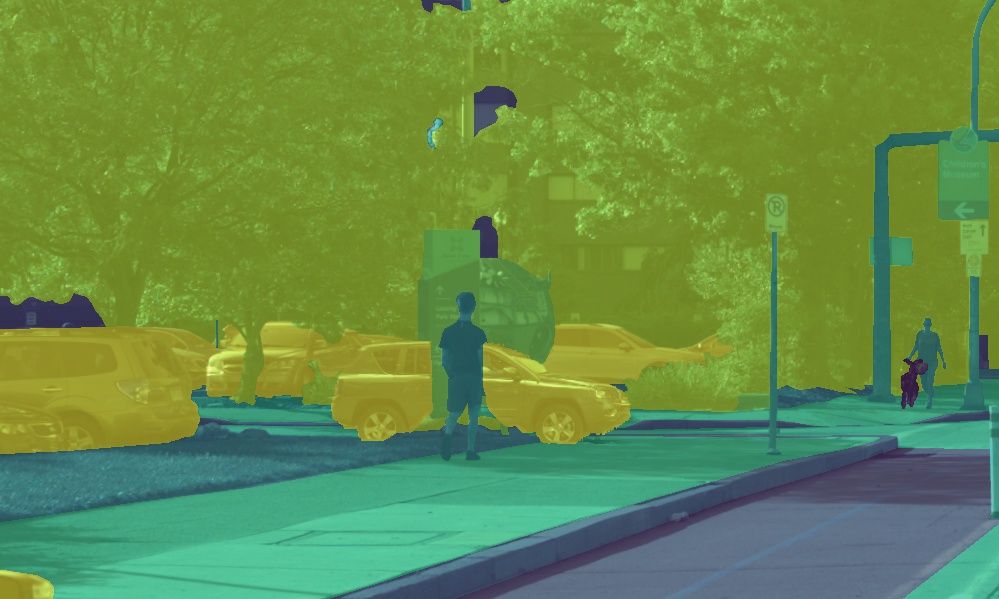}

        \\
        \includegraphics[width=0.33\linewidth,trim={7cm 1cm 8cm  1cm},clip]{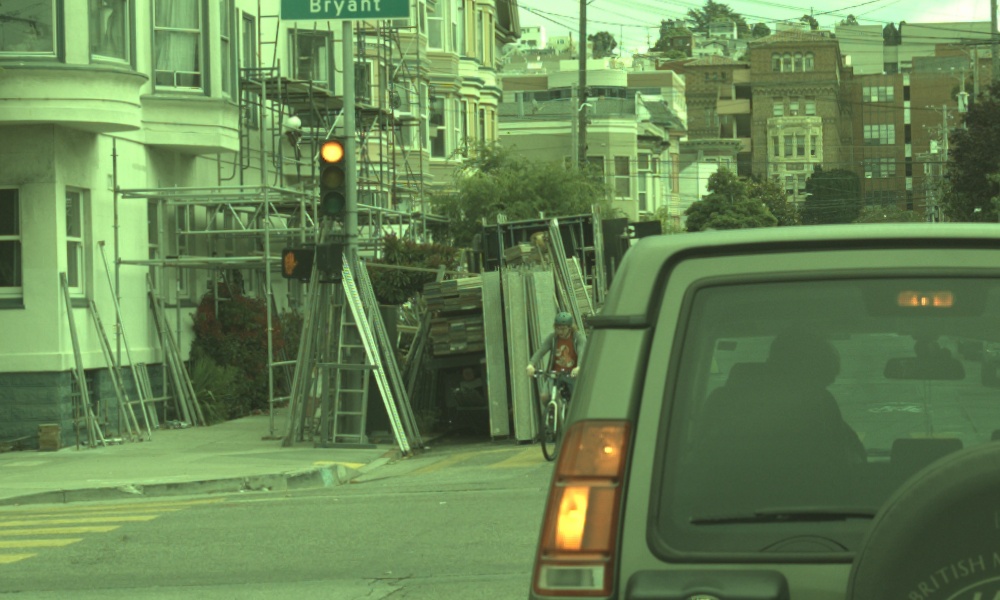}
        & \includegraphics[width=0.33\linewidth,trim={7cm 1cm 8cm  1cm},clip]{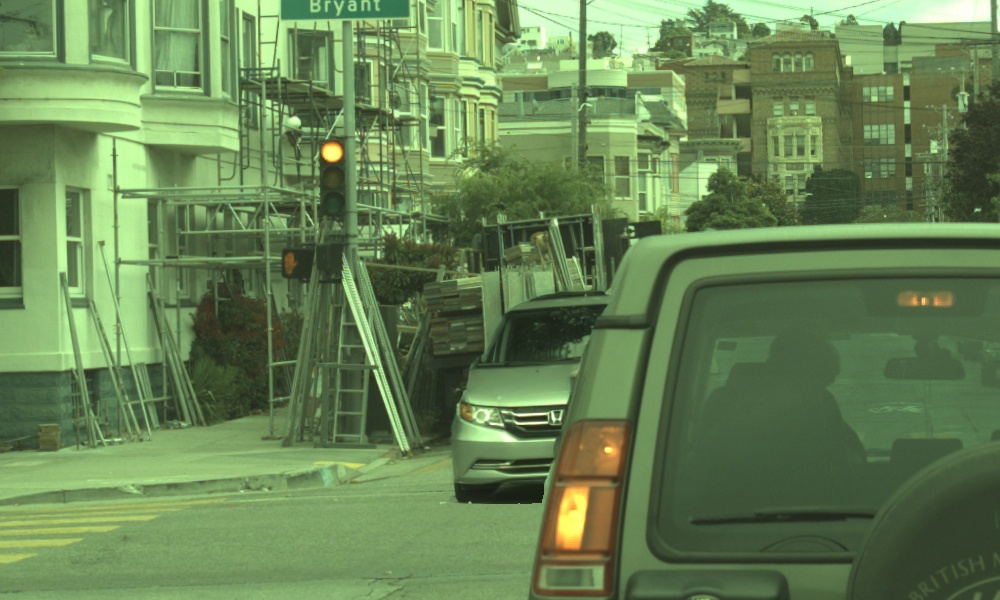}
        & \includegraphics[width=0.33\linewidth,trim={7cm 1cm 8cm  1cm},clip]{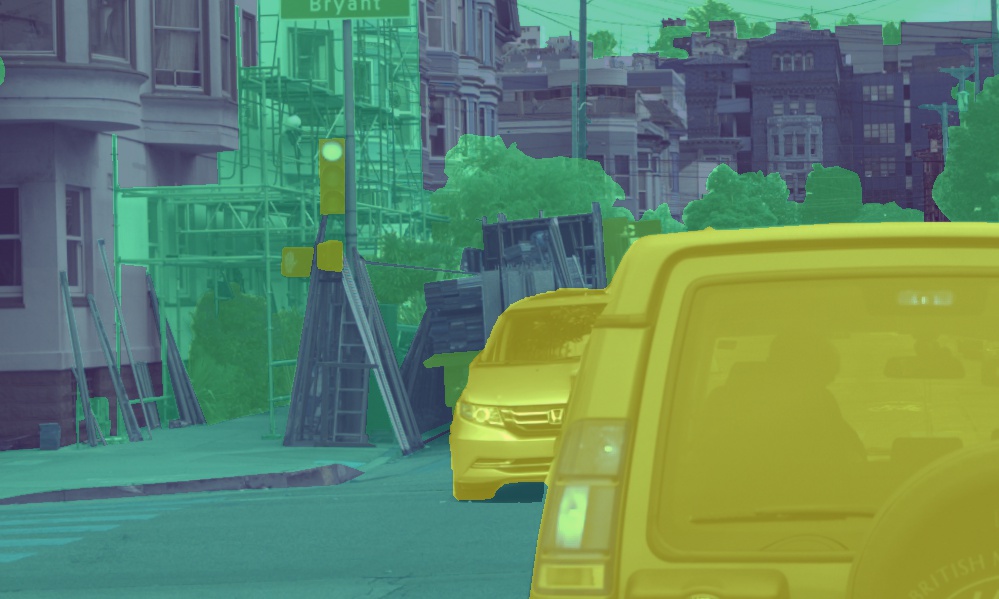}

        \\
    \end{tabular}
    \cutcaptionup
    \caption{\textbf{GeoSim for data augmentation.} Left: image before augmentation. Middle: image after augmentation. Right: augmented  semantic \yun{annotation}.}
    \cutcaptiondown
    \vspace{-3mm}
    \label{fig:dataug}
    \end{figure}

\cutsubsectionup
\subsection{Downstream Perception Task}
\cutsubsectionup
\label{sec:exp_real2sim}

We now investigate
data augmentation, where we use 
labeled real data combined with GeoSim to get
performance gains, without the cost of large scale annotations (as seen in Fig.~\ref{fig:dataug}).}
We first train a segmentation model on
 labeled real data with around 2000 images.
We then use GeoSim to augment these images with inserted vehicles, obtaining 9879 additional training examples in total.
We re-train the segmentation model on both real and augmented data for the same number of iterations.
We evaluate the performance on real data and report the results on Tab.~\ref{tab:sim2real}. 
With these additional training images, we can further boost perception performance by $3.4\%$ (or $0.4\%$ on DeepLabv3~\cite{chen2017rethinking}) for car category and $1.8\%$ (or $0.2\%$ on DeepLabv3~\cite{chen2017rethinking}) for overall mIOU on PSPNet~\cite{zhao2017pspnet}. Importantly we can show consistent improvements across two segmentation models. %

\cutsectionup
\section{Conclusion}
\cutsectiondown
In this work we presented a novel geometry-guided simulation procedure for flexible generation and rendering of synthetic scenes. 
Not only is our approach the first of its kind to fully take into consideration physical realism for dynamic object placement into images, it also bypasses the need for manual 3D asset creation and achieves greater visual realism than competing alternatives. %
 Moreover, we demonstrated improvements in downstream tasks through applications of our technique to 
 \yun{semantic segmentation}.
 There are many exciting follow-up directions opened up by this work such as sim2real, autonomous system 
 evaluation, video editing, {etc}. %
 and we look forward to future extensions of \shorttitle{}.

{\small
\bibliographystyle{ieee_fullname}
\bibliography{references}
}

\appendix
\onecolumn

{\hspace{-5mm} \LARGE{\textbf{Appendix}\\
}}

\maketitle

\textit{  In this supplementary material, we first describe additional technical details on  3D reconstruction, semantic retrieval, depth completion, post-composition,  video simulation, dataset breakdown as well as  A/B human test design (Sec. 1). Furthermore, we provide an extensive collection of qualitative results on comparisons, qualitative ablation on different modules of post-synthesis, and failure case analysis  (Sec. 2). Finally, we recommend the reviewers to watch our supplementary video ($supp\_video.mp4$), which contains an overview of our approach and video simulation results. 
}

  
\section{Additional Technical Details}

\subsection{3D Reconstruction Network}

\begin{figure*}[t]
	\centering
	\includegraphics[width=0.8\textwidth]{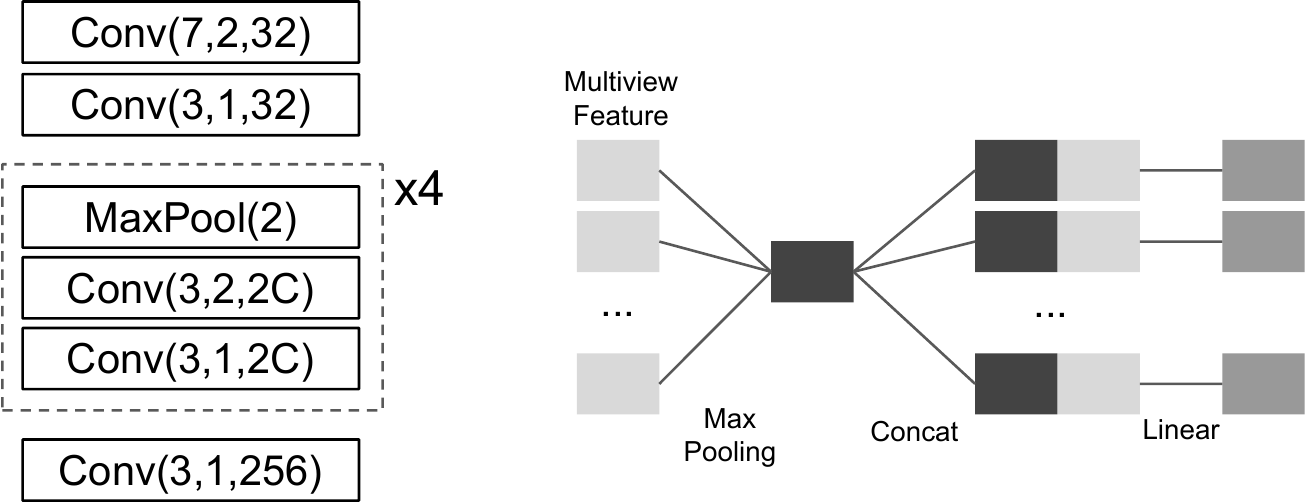} \\
	\caption{\textbf{3D reconstruction network architecture.}
	 Left: Image feature extraction backbone; Right: Multi-view image fusion block.
	\label{fig:meshnet_arch}}
	\end{figure*}

Our 3D reconstruction network takes cropped images and LiDAR sweeps from multiple viewpoints. All cropped images are padded to have the same height and width and are then resized to 256 $\times$ 256.  A small fully convolution network (as seen in Fig.~\ref{fig:meshnet_arch}) is used to extract image features. Note that in the figure, $Conv(K,S,C)$ refers to a convolution layer with kernel size $K$, stride $S$ and output channel $C$. Padding is adjusted to make sure the output size is the same as the input. GroupNorm~\cite{wu2018group} with 32 channels per group is used after each convolution. ReLU is used as the non-linear activation. %
The final output is flattened to a single feature vector for each image. 

To fuse the image features from multiple views, we design a fuse block (as shown on the right of Fig \ref{fig:meshnet_arch}). In the block, multiview features are maxpooled to a single vector, which is then concatenated with each input. The augmented feature vector for each view is processed by a two layer MLP. The dimensions of all hidden layers and output layers of the MLP are 1024. 
We adopt 4 fuse blocks and maxpool to generate the final image-based feature vector. We adopt a standard PointNet~\cite{qi2016pointnet}
as the LiDAR feature extractor, which produces a feature vector with size 1024. Two linear layers consume the final concatenated LiDAR and image features to produce the mean-shape mesh deformation $\delta V$.  The mean-shape mesh is initialized from an icosasphere with 2562 vertices and 5120 triangle faces.

As defined in the paper, we supervise the mesh reconstruction pipeline using silhoutte consistency loss, LiDAR consistency loss and the mesh regularization terms. The \emph{regularizers} are applied to enforce prior knowledge over the resultant 3D shape, namely local smoothness on the vertices as well as normals. 
$$\ell_{\textrm{regularization}}(\mathbf{M}_i) = \alpha \ell_{\textrm{edge}}(\mathbf{M}_i) \;+\; \beta\ell_{\textrm{normal}}(\mathbf{M}_i) \;+\; \gamma \ell_{\textrm{laplacian}}(\mathbf{M}_i)$$
The \emph{edge regularization} term penalizes long edges, thereby preventing isolated vertices.  
$
	\ell_{\textrm{edge}}(\mathbf{M}_i) = %
	{\sum_{\mathbf{v} \in \mathbf{V_i}} \sum_{\mathbf{v}^\prime \in \mathbf{N_v}} \norm{\mathbf{v}-\mathbf{v}^\prime}_2^2}
$,
with $\mathbf{N_v}$  the first ring neighbour vertices of a given vertex $\mathbf{v}$. %
The \emph{Laplacian regularization} \cite{mesh_laplacian} preserves local geometry and prevents intersecting mesh faces by encouraging the centroid of the neighbouring vertices to be close to the vertex:
$
	\ell_{\textrm{laplacian}}(\mathbf{M}_i) = {\sum_{v \in \mathbf{V_i}} %
	\norm{ \sum_{\mathbf{v}^\prime\in \mathbf{N_v}} (\mathbf{v}-\mathbf{v}^\prime)}_2^2}
$.
The \emph{normal regularization} enforces smoothness of the local surface normals, i.e., neighbouring faces are expected to have similar normal direction:
$
	\ell_{\textrm{normal}}(\mathbf{M}_i) = %
	{\sum_{(i, j) \in \mathbf{N}_\mathbf{F}} (1 - \langle \mathbf{n}_{i}, \mathbf{n}_{j}\rangle)}
$,
with $\mathbf{N}_\mathbf{F}$  the set of all the neighbouring faces indices, and $\mathbf{n}_i$ the surface normal of a given face $\mathbf{f}_i$.

We set $\alpha, \beta, \gamma$ to 10.0 in our experiments.
The model is trained for 200 epoches  with 4 input views and batch-size 64 on 16 GPUs. We train it using Adam optimizer with initial learning rate 0.001 and decaied by 0.1 at 150, 180 epoch respectively. It takes about 6 hours to train.

\subsection{Segment Retrieval Details}

\paragraph{Single-view Segment Retrieval:}
To retrieve object-views for rendering in target-view, we first eliminate candidates from significantly different viewpoints (larger than 10\si{\degree}  view changes to the target view).  Then we rank the existing objects by considering their similarity in relative view angle $\theta$ and distance $d$ from the camera.

\[ \textrm{score}(\textrm{object}_\textrm{tgt}, \textrm{object}_\textrm{src}) = \abs{\theta_\textrm{tgt}-\theta_\textrm{src}} + 5\cdot\max(d_\textrm{tgt}-d_\textrm{src},0) \]

Objects are then sampled (as opposed to a hard max) according to a categorical distribution weighted by their inverse score.

\paragraph{Multi-view Segment Retrieval:}

To retrieve  object  with multi views for rendering in videos and multi-cameras, we consider the view range of source object. For every object, we first calculate the view ranges overlap with the target object $\Delta \Theta$, and filter out source objects with  small overlap ($\Delta \Theta<20 \si{\degree}$). Then we rank the existing objects by considering their overlap and minimum distance $d_\textrm{src}$ from the camera.
\[ \textrm{score}(\textrm{object}_\textrm{tgt}, \textrm{object}_\textrm{src}) = 2\cdot { \Delta \Theta} + 5\cdot\max( \min{ d_\textrm{tgt}}- \min{d_\textrm{src}},0) \]
Objects are then sampled (as opposed to a hard max) according to a categorical distribution weighted by their inverse score.

  \subsection{Depth Completion}
 
To realistically place the simulated object in the new scene, we need to infer the occlusion relations between the simulated object and the existing scene elements. To do that, we compare the rendered depth of each simulated object's pixel with the corresponding pixel of the background scene. Since the partial LiDAR sweep belonging to the background scene is not dense enough, we first perform depth completion to generate a dense depth map for the corresponding scene.
In this section, we first provide the ground-truth dense depth dataset preparation details, followed by the architectural details of the depth completion model.

\paragraph{Training Data Preparation:}
 The UrbanData dataset has long trajectories of LiDAR sensor and camera sensor data with manually annotated detection labels. We first generate a dense LiDAR point cloud by aggregate the multi-sweep LiDAR sensor data. For the static background scene, we aggregate the multi-sweep data by compensating the ego-motion of the SDV. For all the objects (filtered out using the detection labels), we aggregate the multi-sweep data by transforming each of them to the corresponding object-coordinate system. For the dynamic objects, we additionally perform color-based ICP to better register the multi-sweep data. We further densify all the aggregated ``vehicle'' point clouds by converting each of them to a dense watertight mesh using a pre-trained implicit surface reconstruction model, DeepSDF \cite{park2019deepsdf}.  The "non-vehilce" categories were densified by splatting each LiDAR point onto a triangular surfel disk. 
 The aggregated point cloud is then rendered to the corresponding camera images to generate the dense depth maps. We first render the background scene aggregated points, followed by (instance-segmentation map aware) rendering of all the detected objects in ascending order of the objects median depth. We used manually annotated ground-truth instance-segmentation maps for depth map refinement.

\paragraph{Architecture Details:}
We use the generated dense-depth dataset to train a depth completion model.
For the model, we use same network architecture as DeepLabV3 \cite{chen2017rethinking}, except the first and the last convolution layers. The input to the model is a concatenated array of camera image, projected sparse depth image, projected sparse depth mask, and the dilated (dilated to 9-neighbouring pixels) sparse depth map. We intialize the DeepLabV3 architecture with the pre-trained COCO weights.

\subsection{Shadow Generation Details}

\begin{figure*}[t]
    \centering
    \includegraphics[width=\linewidth]{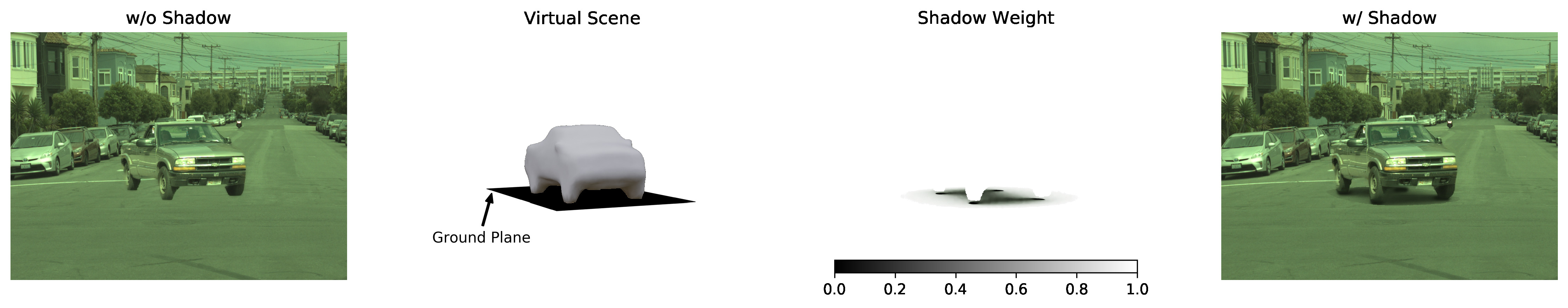}
    \caption{\textbf{Schematics of shadow generation}. (left to right): result without shadow, schematics of virtual scene, shadow weight (ratio of intensity between rendered image with inserted object and without inserted object), result with shadow}

    \label{fig:shadow-gen-supp}
    \end{figure*}

As we discussed in Sec.~4.2 in the paper, Fig.~\ref{fig:shadow-gen-supp} shows the procedure that we applied to generate shadows.
More  qualitative comparison results on the difference between whether applied shadow generations are shown in Fig.~\ref{fig:qual-baselines-supp}.

  \subsection{Post-composition Synthesis Details}

\begin{figure}[t]
\centering
\includegraphics[width=1\textwidth]{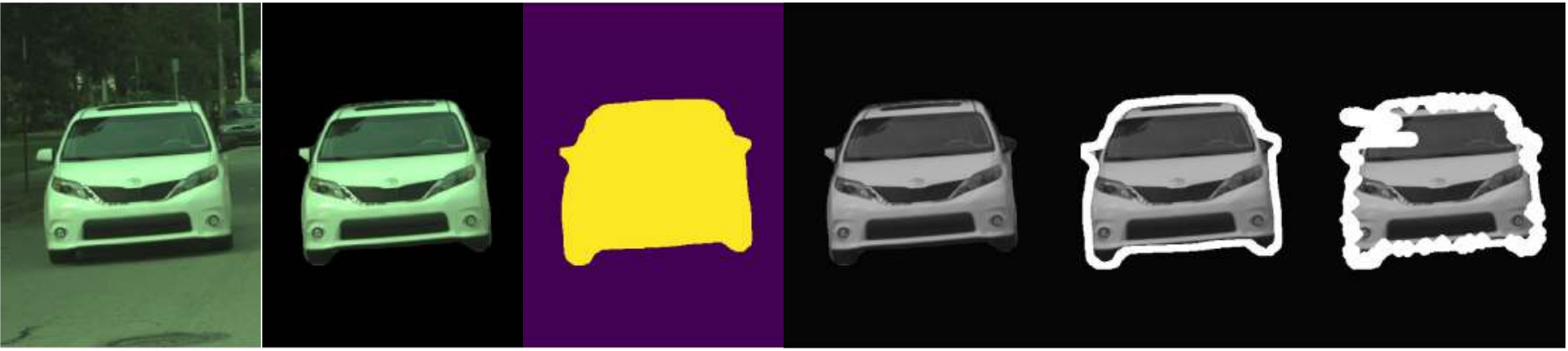} \\
\caption{\textbf{Input data preparation for training the synthesis network}. From left to right: scene image $I$, object segment $S$ and mask $M$ and three random data augmentation including color-jitter, segment boundaries erosion-expansion and random mask in the boundary.
\label{fig:data_aug}}
\end{figure}

\paragraph{Training Data Preparation:}
Our synthesis network is trained on dynamic object images with per-pixel instance labels inferred by \cite{kirillov2020pointrend} in the target scene.
 Given a scene image $I$, we first sample a vehicle binary mask $M$ in that scene, as well as its corresponding RGB segment $S=I\cdot M$.  
Then we apply data augmentation on the segment and mask to mimic the noisy input at the inference stage, with color inconsistency, missing texture and imperfect boundary.  Specifically, we applied 1) random color jitter on the scene and segment separately, which randomly changes the brightness, contrast and saturation to mimic color inconsistency between foreground and target scene;  2) random erosion on the segment boundary from 3 to 20 pixels and a random dilation on the mask $M$ from 3 to 40 pixels to blend object boundary naturally; 3) a random drop on the segment $S$ with 0.1\%--1\% of total boundary pixels and applied random dilation on those samples to mimic missing textures of the inserted virtual object. Please refer to Fig.~\ref{fig:data_aug} for illustration of such data augmentation process. 

\paragraph{Architecture Details:}
Our synthesis network architecture is inspired by \cite{yu_free-form_2019}.  One difference is that our network takes the instance segment mask as an additional input. Thus, the network takes as input the object segment, target scene and mask region.  We crop a 512$\times$512 region centered around the object center from the full scene. 

\paragraph{Loss Functions:}
We apply perceptual loss \cite{DBLP:journals/corr/JohnsonAL16} on the generated image $I'$ and $I$, using the \texttt{conv3\_3} feature activations in a pretrained VGG16 network $F_v$.
\[ L_G^{\textrm{perc}}  = \sum \norm {F_v(I) - F_v(G(I\cdot (1-M^A), M^A, S^A))}_1 \]
A GAN loss is also applied to optimize the synthesis network.
\[ L_G^{\textrm{gan}} = - \mathbb{E}_{z\sim P_z(z)}[D(G(I\cdot (1-M^A), M^A, S^A))] \]
For the discriminator, we adopt the same loss as \cite{yu_free-form_2019}.

\paragraph{Inference:}
At the inference stage during simulation, our network takes the raw composition image from novel view warping and occlusion reasoning as input and produces natural blended results. Specifically, we crop a square region with the visible rendered segment in the centre, which is twice the size of the larger side of rendered vehicles ($size$). The cropped size is set to be $128\times 128$ at least and  $1024\times 1024$ at most.  Invisible pixels due to inverse warping are filled with zeros. {Then we c erode the rendered segment by $\dfrac{size}{64}$ pixels and dilate the mask by $\dfrac{size}{32}$ pixels}. These inputs are fed into the synthesis network and the final outputs paste back to the original location.

\subsection{Video Simulation Details}

In order to simulate a realistic video footage with new dynamic objects inserted, we first select a subset of snippets from our dataset of real-world logs, each consisting of 50 consecutive frames sampled at 10\si{\hertz} (for a total duration of 5s) to augment with up to 5 simulated objects using our approach. 

For these relatively short sequences, we sample the object placement within the initial camera field-of-view in the first frame of the sequence and equip the object with a realistic path using the local lane graph, as illustrated in Figure 3 of the main text.  

To retrieve an object for insertion from the 3D asset bank, we consider the view-range of each source object. For every object, we first calculate the view-range overlap with the target object, $\Delta \Theta$, and filter out source objects with small overlap ($\Delta \Theta<20 \si{\degree}$). Then we rank the existing objects by considering their overlap and minimum distance $d_\textrm{src}$ from the camera.
\[ \textrm{score}(\textrm{object}_\textrm{tgt}, \textrm{object}_\textrm{src}) = 2\cdot { \Delta \Theta} + 5\cdot\max( \min{ d_\textrm{tgt}}- \min{d_\textrm{src}},0) \]
Objects are then sampled (as opposed to a hard max) according to a categorical distribution weighted by their inverse score.

We now describe how we use a set of heuristics-based behavior and lateral/longitudinal driving models to refine the path into a timestep-by-timestep \emph{trajectory} of \emph{kinematic states} comprising of location, orientation, velocity, and acceleration. This realistic trajectory simulation helps the added vehicle in the simulated video achieve realistic motion such as braking and acceleration. 
After executing segment retrieval and performing a final collision check, the object and its refined trajectory is then inserted into the log for inclusion. Additional objects can be added in a similar fashion.

We use a cost-based heuristic behavior model for determining lane change actions, a heuristic lateral model for the side-to-side motion of the object, and the Intelligent Driver Model~\cite{schulz_interaction-aware_2018}%
for the longitudinal movement. We define the following notation: %
\begin{align*}
	p_x &= \textrm{longitudinal position} \\
	v_x &= \textrm{longitudinal velocity} \\
	p_y &= \textrm{lateral position} \\
	v_y &= \textrm{lateral velocity}.
\end{align*}
All models use a frequency, or reaction time, of 10\si{\hertz}. 

\subparagraph{Behavior Model Details:} Two actors are considered as colliding if the inter-vehicle distance between them is less than 2m. Let $d_h,d_f$ denote the distances between the given vehicle and the headway (nearest front) vehicle or following (nearest back) vehicle respectively. Similarly, define $c_h,c_f$ to be the headway and following costs. The cost functions we use are: 
\begin{align*}
	c_h &= 
	\begin{cases}
		10^8 & d_h < 2 \\
		\tfrac{10^4}{d_h} & \textrm{otherwise}
	\end{cases} \\
	c_f &= 
	\begin{cases}
		10^8 & d_f < 2 \\
		\tfrac{10^2}{d_f} & \textrm{otherwise}.
	\end{cases}
\end{align*}
The cost of making a lane change is $10^3$, and the cost $c_l$ for being close to the end of a lane when distance $d_l$ from the lane end is 
\[ 
c_l = \tfrac{10^5}{d_l}.
\]

\subparagraph{Lateral Model Details:} We use a simple heuristic for the target lateral speed that seeks to return to the lane centerline, bounded by a function of the longitudinal (forward-backward) movement. Specifically, 
\[ v_x = \min(-p_x, 0.1 v_y), \]
clipped so that the maximum acceleration magnitude is 3\si{ms^{-2}}. 

\subparagraph{Longitudinal Model Details:} We use the following parameters: minimum and maximum target speeds of 15\si{ms^{-1}} and 25\si{ms^{-1}} respectively, acceleration exponent of 4\si{ms^{-2}}, maximum acceleration and deceleration magnitudes of 5\si{ms^{-2}}, minimum gap of 2\si{m}, headway time of $1.5$\si{s}, and default vehicle length of $4.5$\si{m}.

\subsection{UrbanData Dataset breakdown}

In the UrbanData, we have roughly 16.5K labelled snippets. Out of these, 12K snippets have manually annotated 2D segmentation maps (one frame per snippet). As shown in Tab.~\ref{tab:urban-dataset-breakdown}, we divide these 16.5K snippets into multiple splits for training and testing various components of the GeoSim pipeline as well as the baslines.
Tab.~\ref{tab:urban-dataset-experiment-breakdown} maps each task to its corresponding training/ testing dataset split.
Note that GeoSim does not need any gt labeled data. All labeled data are for training/evaluation purposes.

\begin{figure}[t]
\begin{minipage}[c]{0.3\textwidth}
\resizebox{0.95\textwidth}{!}{%
\setlength{\tabcolsep}{3pt}
\begin{tabular}{lcccc}
\specialrule{.2em}{.1em}{.1em}
Dataset Split & \#Logs & \#Frames used \\
\hline
Split A & 5K & 1.2M \\
Split B &  7K & 7K \\
Split C & 2.8K & 2.8K \\
Split D & 2K & 2K \\
\specialrule{.2em}{.1em}{.1em}
\end{tabular}
}
\captionof{table}{\textbf{UrbanData dataset splits.}}
\label{tab:urban-dataset-breakdown}
\end{minipage}
\hfill
\begin{minipage}[c]{0.65\textwidth}
\resizebox{0.95\textwidth}{!}{%
\setlength{\tabcolsep}{3pt}
\begin{tabular}{lcccc}
\specialrule{.2em}{.1em}{.1em}
Task & SubTask & Dataset Split  & Label Used\\
\specialrule{.1em}{.05em}{.05em}
GeoSim & Mesh Reconstruction training & Split A & 3D Box and mask from \cite{kirillov2020pointrend} \\
GeoSim & Post-Compostion training & Split A &  mask from \cite{kirillov2020pointrend} \\
GeoSim & Depth Completion training & Split B & Aggregated Lidar \\
GeoSim & Whole pipeline & Split C & 3D Box from \cite{liang2020pnpnet} \\
\hline
Baselines &  Training & Split B & GT Semantic Mask \\
Baselines & Test & Split C  & GT Semantic Mask \\
Downstream & Sim2Real Training & Split C & GT Semantic mask and GeoSim Mask \\
Downstream & Sim2Real Evaluation & Split D  & GT Semantic mask\\
\specialrule{.1em}{.05em}{.05em}
\end{tabular}
}
\captionof{table}{\textbf{UrbanData experimental setting.}}
\label{tab:urban-dataset-experiment-breakdown}
\end{minipage}
\end{figure}

\subsection{A/B interface and instructions}

\begin{figure*}[t]
\includegraphics[width=\linewidth,trim={0 3cm 0 0},clip]{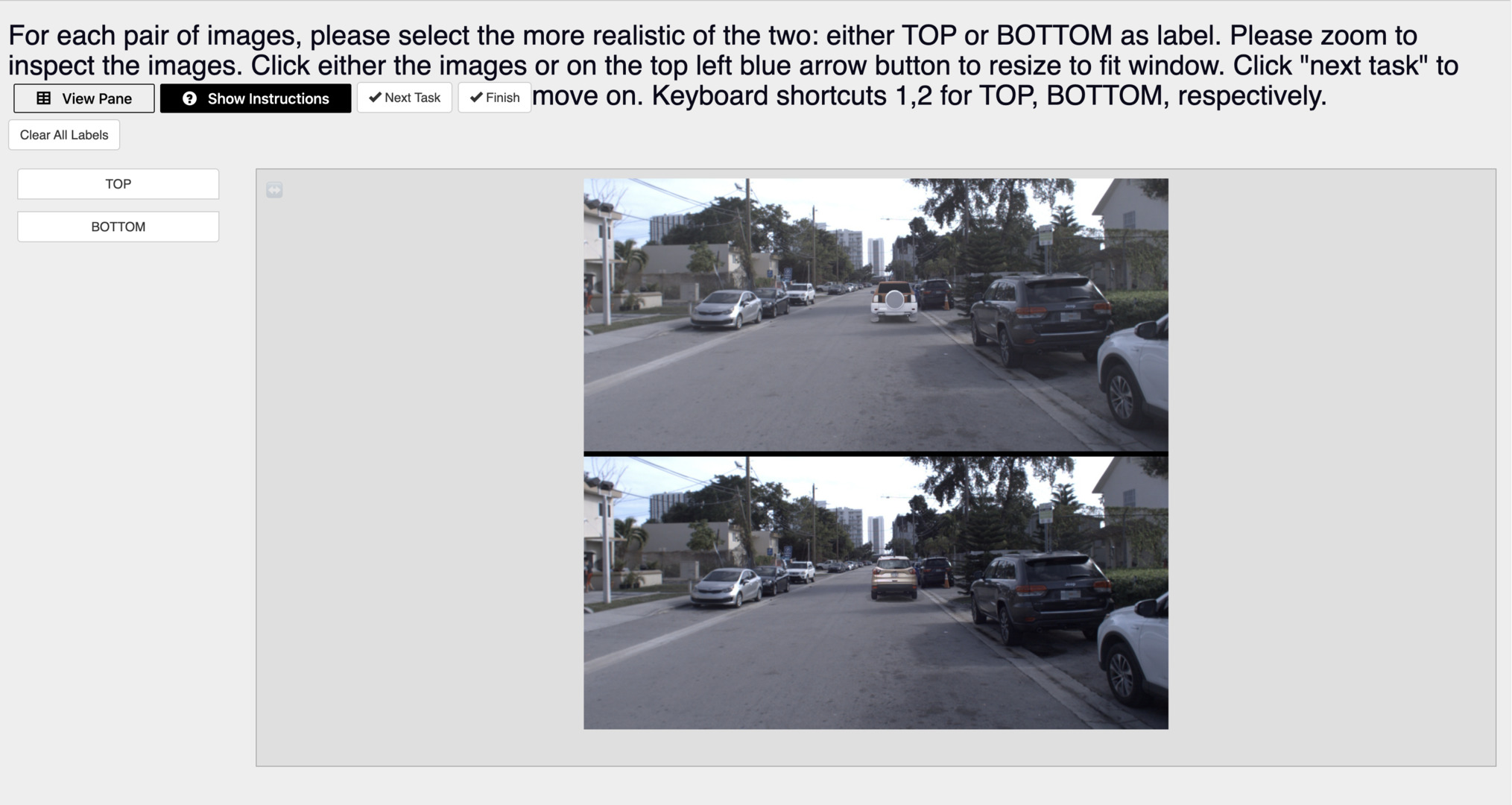}
\vspace{-1em}
\caption{\textbf{A/B test user interface}. Users must select which image ("TOP" or "BOTTOM") they consider most realistic. 
\label{fig:ab_interface}}
\end{figure*}

As discussed in Section 5.2 of the main paper, we performed a human study to demonstrate that GeoSim images appear more realistic compared to other baseline methods. For interface simplicity and ease of annotation, we performed pairwise comparisons instead of ranking. We also perform pair-wise comparison over ranking to mitigate user-bias, as all the baselines we compare against use the same object proposal method. An example interface is shown in  Fig. \ref{fig:ab_interface}. Each user would be provided two images, one generated with GeoSim and one generated with one of the baseline methods. Each image would be assigned with equal probability to the top or bottom location. Both the baseline and GeoSim would receive the same input real image and semantic segmentation. Based on the method and placement procedure, an object would be added to the scene,  not necessarily in the same location (as seen in Fig. \ref{fig:ab_interface}). 
We asked 16 users who are familiar with self-driving but who had limited or no knowledge of image simulation or of our method to select which image they prefer. 
On average, users annotated approximately 120 images, for a total of $\sim$1900 images.  
Here are the detailed instructions we provided along with each query:
Detailed instructions:
\begin{quote}
For each pair of images, please select the more realistic of the two by selecting either {TOP or BOTTOM} 
as the image label. Make sure to consider all relevant aspects of realism including but not limited to: visual appearance, lighting, shadows, relationship between elements within the image (ie occlusions), consistency in color, weather conditions, positioning on the road, and traffic regulations. We recommend clicking either on the images or on the top left blue arrow button to resize both images to fit the window. Click "next task" to move on. Note you will not be able to go back to a previous image, and the "Finish" button has no effect until all examples have been labeled, so there is no risk in accidentally clicking it. You can use the keyboard shortcuts 1,2 for TOP, BOTTOM, respectively.
\end{quote}
{
We compute the success rate, i.e. whether our method is preferred over a specific baseline as: $\frac{\textrm{\# of times GeoSim selected}}{\textrm{\# of pairs with GeoSim and baseline}}$.
{We perform one-tailed binomial testing with the null hypothesis that GeoSim is not better than the baseline in over 50\% of cases.
 The maximum $p$-value across all human evaluations} %
 is 1.64e-18 (when GeoSim was preferred in 211/279 pairs in the 2D synthesis ablation), indicating statistical significance.
}

  
\section{Qualitative Results}

\subsection{Visual Comparisons}
\begin{figure*}[t]
    \centering
    \includegraphics[width=1\textwidth]{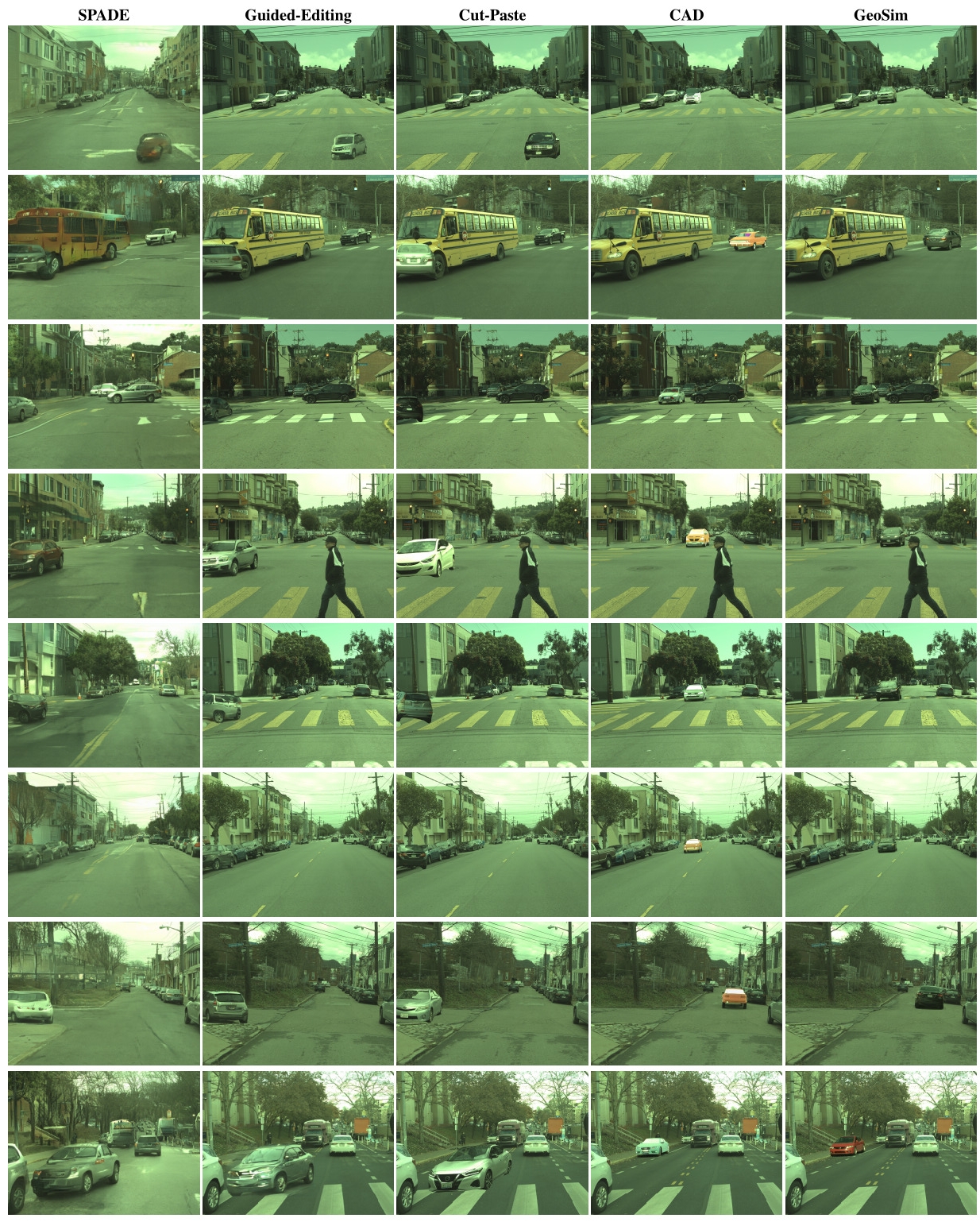} \\
    \caption{\textbf{Qualitative Comparison of Image Simulation approaches.}}
    \cutcaptiondown
\label{fig:qual-baselines-supp}

    \end{figure*}

\begin{figure*}[!ht]
    \centering
\includegraphics[width=1\textwidth]{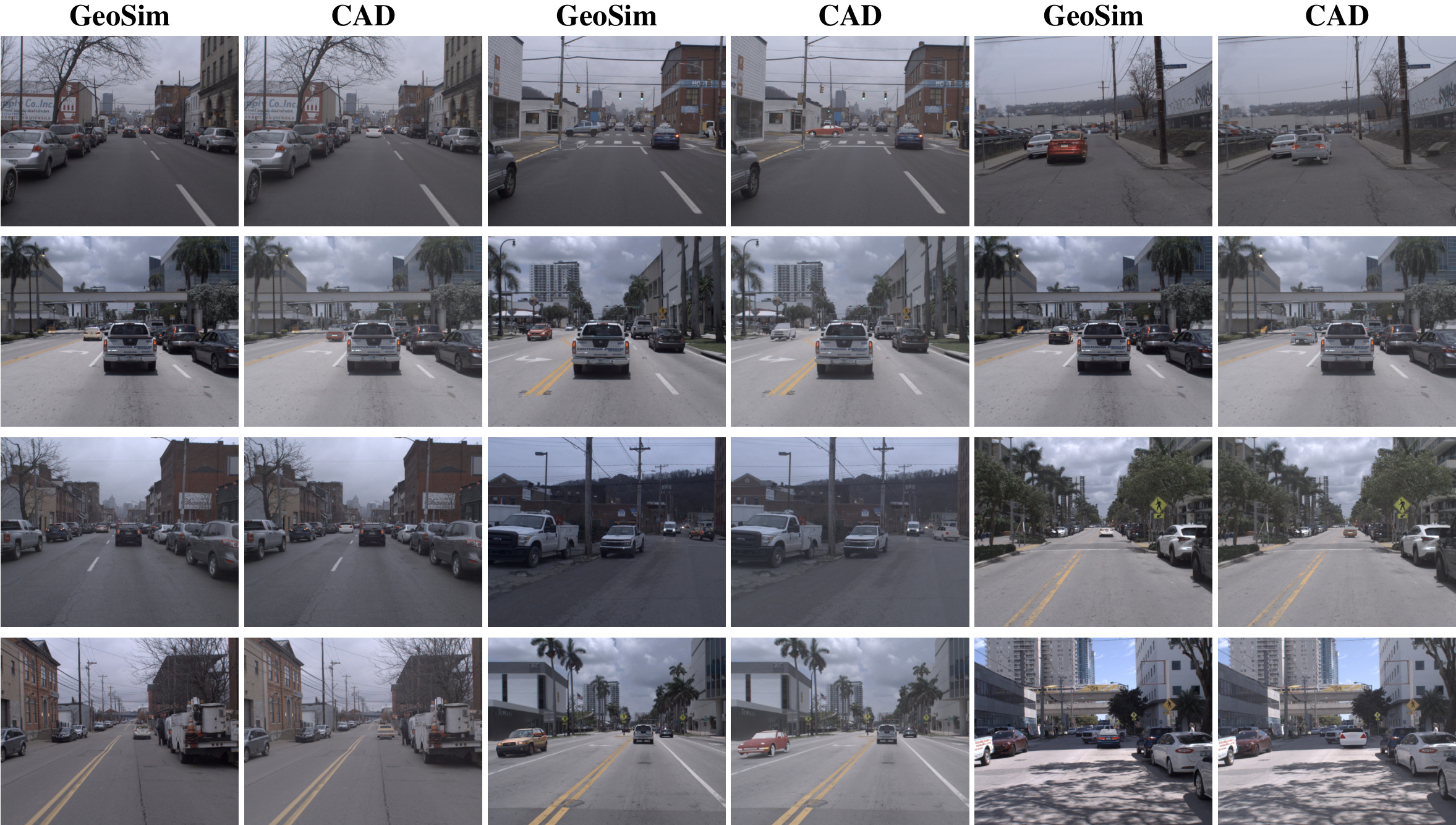} \\
\caption{\textbf{Qualitative Comparison of Image Simulation approaches on Argoverse.}}
\label{fig:argo-qual-results-supp}
\end{figure*}

In addition to the qualitative results shown in Fig.~5 in the paper, we further showcase more qualitative comparisons among the previously-discussed image simulation baselines in Fig.~\ref{fig:qual-baselines-supp}. As can be seen from Fig.~\ref{fig:qual-baselines-supp}, GeoSim produces much more realistic and 3D aware simulated images, compared to the visually significant failures (e.g., blurred textures, implausible placements, distorted shapes, boundary artifacts)  produced by the other simulation methods. We also showcase more qualitative comparisons on public dataset Argoverse in Fig.~\ref{fig:argo-qual-results-supp}.

\subsection{Ablation Analysis}

We also conduct qualitative ablation on three key components in post processing: occlusion reasoning, shadow and synnet. We compare GeoSim results against the one with the selected component removed. As seen in Fig. \ref{fig:qual-abl}, with any one of the component being removed, the realism of the synthesis results drop significantly. 
Removing occlusion reasoning, the synthesized vehicles aren't able to conform with the existing scene elements. Removing Shadow, the synthesized vehicles seems to be off-the ground. Removing synnet, the synthesized vehicles show inconsistent illumination and color-balancing w.r.t the target scene and discrepancies at the boundaries. Please zoom in to see the detail.

{
In addition, we also show lane map and synthesis network ablations.
Specifically, Fig.~\ref{fig:lanemap}
shows results of GeoSim using random sampling instead of the lane map, 
where we uniformly sample empty ground locations according to the LiDAR sensor data 
and draw uniform orientations. 
Random sampling generates some interesting and useful edge cases, 
but it is not temporally consistent, making it not amenable to video simulation. 
The A/B test result shows that humans prefer our GeoSim at 95.5\% of the time. 
As for human scores %
for GeoSim without the image synthesis network,
users overwhelmingly  (95.0\%)  prefer the full GeoSim.
}

\subsection{Failure Cases}

\begin{figure*}[ht]
    \centering
    \includegraphics[width=0.95\textwidth]{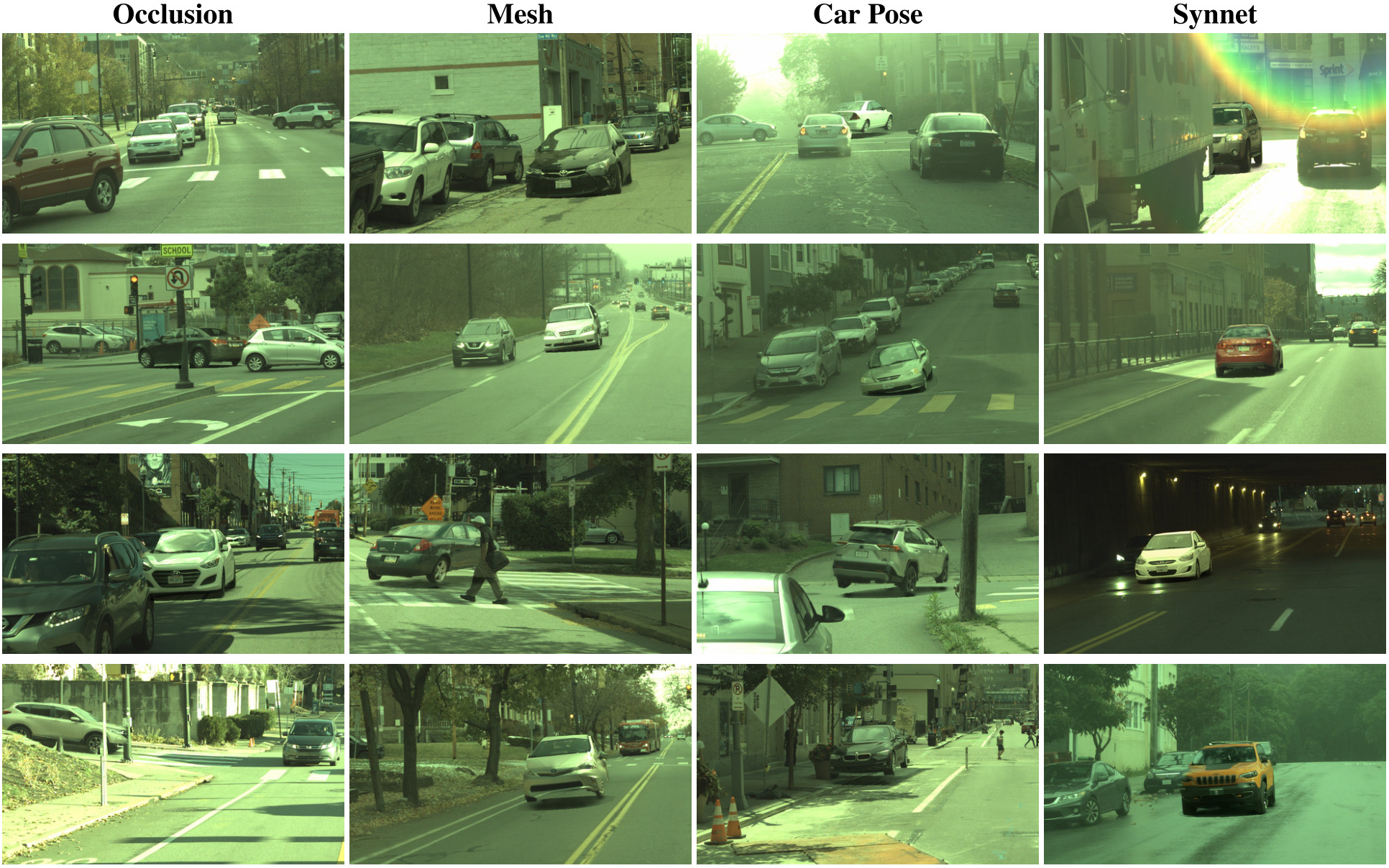} \\

\caption{\textbf{Qualitative visualization of the failure cases.}}
\label{fig:qual-failures-supp}
\end{figure*}

While GeoSim simulated images manages to produces realistic results in many cases,
there is still space for potential improvements. In Fig.~\ref{fig:qual-failures-supp}, we highlight four major failure cases: (1) Incorrect Occlusion Relationships in complicated scene, (2) Irregular reconstructed mesh, (3) Inaccurate object poses, usually caused by Map error and (4) Illumination failure by distinct illumination difference between rendered segment and target scene.

\begin{figure*}[!ht]
    \centering
    \includegraphics[width=1\textwidth]{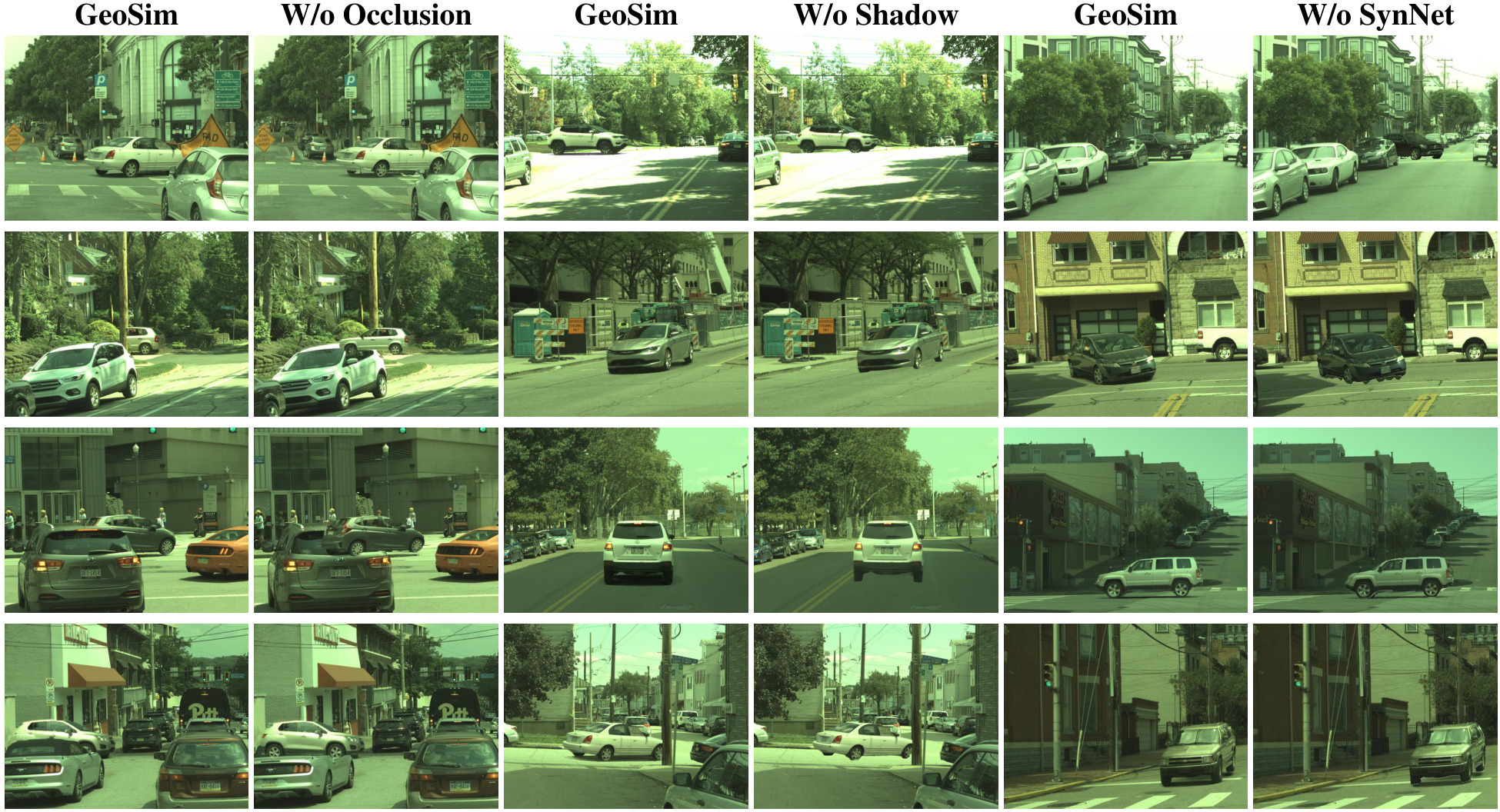} \\
    
    \caption{\textbf{Qualitative ablation on the composition.}}
    \label{fig:qual-abl}
    \end{figure*}

\begin{figure*}[t]
    \centering
    \includegraphics[width=\linewidth]{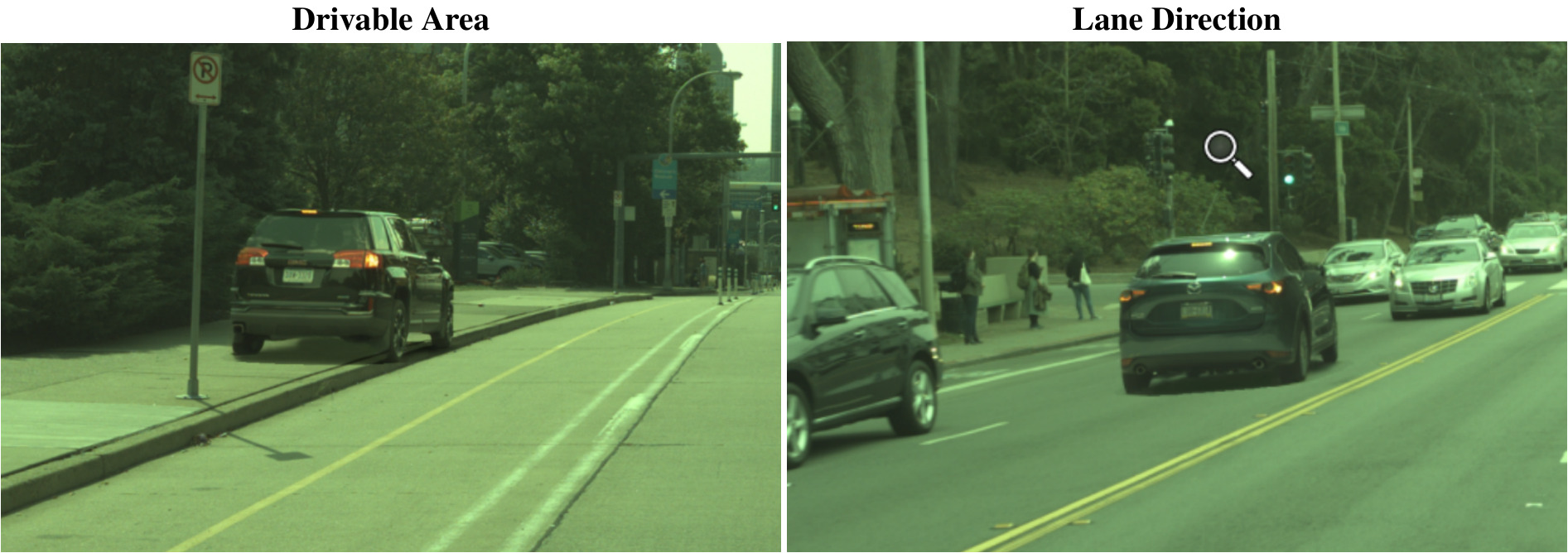}
    \caption{\textbf{GeoSim results without lane map.}}
    \label{fig:lanemap}
    \vspace{-2mm}
    \end{figure*}

  \clearpage



  

\end{document}